%% file: main.tex
\documentclass[11pt, a4paper, logo, copyright, nonumbering]{mlins_template}
\usepackage[numbers,sort&compress,round,comma]{natbib}
\usepackage{dblfloatfix}
\usepackage{ulem}
\usepackage{caption}
\usepackage{dramatist}
\usepackage{xspace}
\usepackage{pifont}
\usepackage{multirow}
\usepackage{tcolorbox}
\usepackage{xltabular}
\usepackage{longtable}
\usepackage{hyperref}
\usepackage{amsfonts}
\usepackage{amsmath}
\usepackage{amssymb}
\usepackage{lineno}
\usepackage{multirow}
\usepackage{adjustbox}

\usepackage[bottom]{footmisc}

\usepackage{CJKutf8}
\usepackage{subfigure}
\usepackage{setspace}

\usepackage{dsfont}
\usepackage{array} %
\usepackage{tabularx} %
\usepackage{subfigure} %
\usepackage{xcolor} %
\usepackage{tabularx}
\usepackage{booktabs}
\usepackage{algorithm}
\usepackage{listings}
\usepackage{parskip}
\usepackage{lipsum}  %
\usepackage{multicol} %
\usepackage{lineno}
\usepackage{soul}

\makeatletter
\def\@BTrule[#1]{%
  \ifx\longtable\undefined
    \let\@BTswitch\@BTnormal
  \else\ifx\hline\LT@hline\textbf{}
    \nobreak
    \let\@BTswitch\@BLTrule
  \else
     \let\@BTswitch\@BTnormal
  \fi\fi
  \global\@thisrulewidth=#1\relax
  \ifnum\@thisruleclass=\tw@\vskip\@aboverulesep\else
  \ifnum\@lastruleclass=\z@\vskip\@aboverulesep\else
  \ifnum\@lastruleclass=\@ne\vskip\doublerulesep\fi\fi\fi
  \@BTswitch}
\makeatother

\addto\extrasenglish{
}

\input{commands.tex}

\renewcommand{\today}{}

\title{\centering Health system learning  achieves generalist neuroimaging models}

\author{
\centering
Akhil Kondepudi\tsc{1,2}\quad
Akshay Rao\tsc{1}\quad
Chenhui Zhao\tsc{1,3}\quad
Yiwei Lyu\tsc{1,3}\quad
Samir Harake\tsc{1}\quad
Soumyanil Banerjee\tsc{1}\quad
Rushikesh Joshi\tsc{1,4}\quad
Anna-Katharina Meissner\tsc{5}\quad
Renly Hou\tsc{1,2}\quad
Cheng Jiang\tsc{1,2}\quad
Asadur Chowdury\tsc{1}\quad
Ashok Srinivasan\tsc{6}\quad
Brian Athey\tsc{2}\quad
Vikas Gulani\tsc{6}\quad
Aditya Pandey\tsc{4}\quad
Honglak Lee\tsc{3}\quad
Todd Hollon\tsc{1,2,3,4}

\small
\tsc{1}Machine Learning in Neurosurgery Lab, University of Michigan\quad 
\tsc{2}University of Michigan Computational Medicine and Bioinformatics\quad
\tsc{3}University of Michigan Computer Science and Engineering\quad 
\tsc{4}University of Michigan Neurosugery\quad
\tsc{5}University of Cologne Neurosugery\quad
\tsc{6}University of Michigan Radiology\quad \\
\href{https://neurovfm.mlins.org/}{neurovfm.mlins.org}}

\begin{abstract}
Frontier artificial intelligence (AI) models, such as OpenAI's GPT-5 \cite{OpenAI2025-qd} and Meta's DINOv3 \cite{Simeoni2025-xo}, have advanced rapidly through training on internet-scale public data, yet such systems lack access to private clinical data. Neuroimaging, in particular, is underrepresented in the public domain due to identifiable facial features within MRI and CT scans, fundamentally restricting model performance in clinical medicine \cite{Ziller2024-nj}. Here, we show that frontier models underperform on neuroimaging tasks and that learning directly from uncurated data generated during routine clinical care at health systems, a paradigm we call health system learning, yields high-performance, generalist neuroimaging models. We introduce NeuroVFM, a visual foundation model trained on 5.24 million clinical MRI and CT volumes using a scalable volumetric joint-embedding predictive architecture. NeuroVFM learns comprehensive representations of brain anatomy and pathology, achieving state-of-the-art performance across multiple clinical tasks, including radiologic diagnosis and report generation. The model exhibits emergent neuroanatomic understanding and interpretable visual grounding of diagnostic findings. When paired with open-source language models through lightweight visual instruction tuning, NeuroVFM generates radiology reports that surpass frontier models in accuracy, clinical triage, and expert preference. Through clinically grounded visual understanding, NeuroVFM reduces hallucinated findings and critical errors, offering safer clinical decision support. These results establish health system learning as a paradigm for building generalist medical AI and provide a scalable framework for clinical foundation models.

\end{abstract}

\begin{document}
\maketitle

\textbf{Keywords:} health system learning, neuroimaging, joint embedding-predictive architectures, artificial intelligence, medical computer vision, foundation models

\clearpage
\input{latex/intro}

\input{latex/results}

\input{latex/discussion}

\input{latex/methods}

\bibliographystyle{unsrtnat}
\bibliography{paperpile.bib,yiwei_additional}

\input{latex/extended_data}

\end{document}

%% file: commands.tex
\newcommand{\tsc}[1]{\textsuperscript{#1}}

\renewcommand{\emph}[1]{\textit{#1}}

\renewcommand{\phi}{\varphi}

\renewcommand{\geq}{\geqslant}

\renewcommand{\epsilon}{\varepsilon}
\renewcommand{\imath}{\mathrm{i}}

\newlength{\restsubwidth}
\newlength{\restsubheight}
\newlength{\restsubmoreheight}
\newcommand{\rest}[2]{%
        \settowidth{\restsubwidth}{\ensuremath{#2}}
        \settoheight{\restsubheight}{\ensuremath{{}_{#2}}}
        \ensuremath{{#1\hskip 0.5pt}_{\vrule\kern2pt\parbox[b][%
        4pt][b]{\the\restsubwidth}{%
                        \ensuremath{{}_{#2}}}}}
        }

%% file: latex/intro.tex
\section*{Main}

\begin{figure*}[t]
    \centering
    \includegraphics[width=\textwidth]{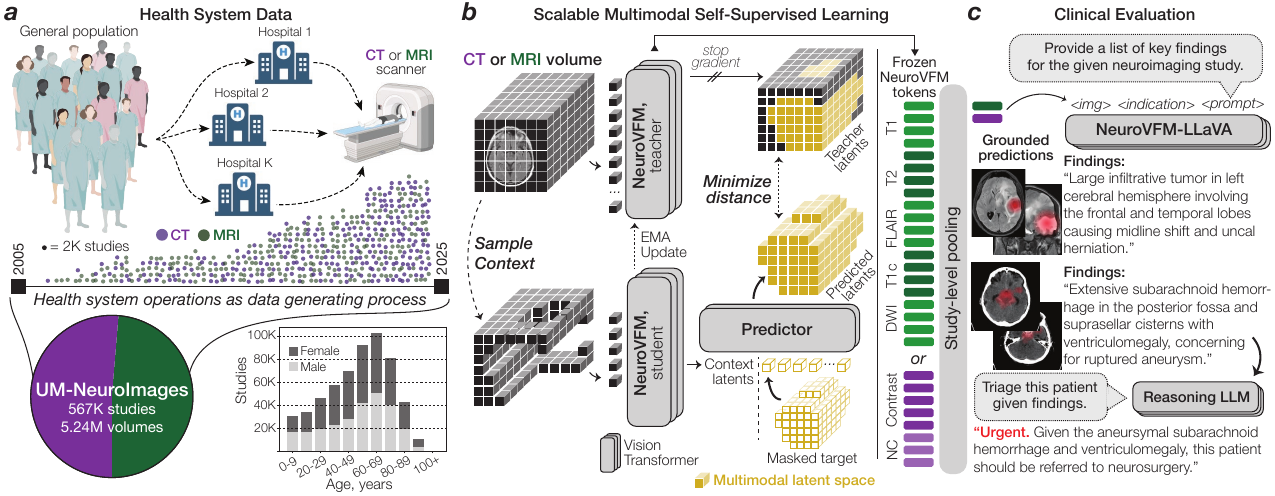} 
    \vspace{-15pt}
    \caption{\footnotesize \textbf{Overview of Health System Learning with NeuroVFM.}
    \textbf{a}, Health system learning directly models the data generating process of clinical operations at large health systems. The UM-NeuroImages dataset comprises 5.24 million volumes from 566,915 studies, acquired over 20 years at Michigan Medicine. Age and sex distribution are shown below.
    \textbf{b}, NeuroVFM was trained using Vol-JEPA, a scalable  volumetric self-supervised method that learns a unified latent space for CT and MRI. A 3D volume is partitioned into a small context and larger masked target with the background removed. The context is encoded by an online (\textit{student}) 3D vision transformer. A predictor combines context latents with position encodings of the masked target region to predict the masked region latents. Ground truth latents for the masked region are generated by an offline (\textit{teacher}) encoder updated by an exponential moving average, with gradients stopped through the teacher. Training minimizes the distance between the predicted and teacher latents using a smooth L1 loss.
    \textbf{c}, At inference, NeuroVFM encodes all volumes in a neuroimaging study into latent visual tokens for downstream tasks. The same visual tokens can be used to fine-tune an open-source multimodal language model (i.e., Qwen3-14B, LLaVA-1.5-style) to generate radiology reports. Illustrative findings and corresponding grounded attention maps are shown. The findings can then be passed to a frontier reasoning LLM (i.e., GPT-5-thinking) for interpretation and triage.}
    \label{fig:fig_1_modeling}
    \vspace{-15pt}
\end{figure*}

Multimodal large language models (MLLMs) derive much of their capability from learning on internet-scale data, enabling these models to approximate the breadth of human experience across language, images, and video. Clinical medicine, however, is underrepresented on the public internet. MLLMs trained exclusively on public data lack access to the rich private information embedded in real-world patient care, which fundamentally limits their performance on clinical tasks. We propose \emph{health system learning} as a new paradigm in which medical foundation models learn directly from uncurated data generated during clinical operations at health systems. By learning in the same complex, nuanced environment in which expert clinicians themselves train, rather than second-hand internet descriptions used by MLLMs, AI models can acquire rich, medical representations grounded in anatomy, pathology, and clinical workflows. MLLMs know the map, health system learners know the territory.

To demonstrate the strength of health system learning, we introduce NeuroVFM, a generalist neuroimaging visual foundation model trained on all clinical MRI and CT studies from a large academic health system. Unlike previous models that rely on data curation, human annotations, or radiology report supervision \cite{Zhang2022-fz, Radford2021-wf}, NeuroVFM is optimized for general neuroimaging through a self-supervised vision-only algorithm called Volumetric Joint-Embedding Predictive Architectures (Vol-JEPA) \cite{LeCun2022-ad}. Our method enforces representation learning across imaging modalities and disease spectra, capturing both global and fine-grained neuroanatomic and pathologic features. Health system learning with Vol-JEPA enables NeuroVFM to achieve state-of-the-art performance, surpassing leading proprietary and open-source frontier models across multiple clinical tasks, including radiologic diagnosis and report generation. NeuroVFM predictions are diagnostically grounded, with pathologic image regions mapped to neurologic diagnoses. When integrated with open-source language models, NeuroVFM acts as a visual understanding module that outperforms GPT-5 and Claude Sonnet 4.5 on neuroimaging interpretation and triage, while reducing hallucinated findings. These results show that NeuroVFM narrows the gap between frontier AI systems and safe clinical deployment.

%% file: latex/results.tex
\subsection*{Learning with Vol-JEPA}
To train NeuroVFM, we assembled a multicenter, health system-scale, multimodal dataset called \textbf{UM-NeuroImages} (Fig. \ref{fig:fig_1_modeling}a). The dataset captures real-world neuroimaging studies ordered by physicians across a diverse patient population as part of routine clinical care. We queried Michigan Medicine's picture archiving and communication system for all MRI and CT imaging studies of the brain, head, neck, face, and orbits, yielding 566,915 studies, 5.24 million 3D volumes, and more than 200 million 2D images. UM-NeuroImages is an order of magnitude larger than previous 3D neuroimaging datasets. A clinically informed diagnostic ontology was defined, encompassing 74 MRI and 82 CT diagnoses spanning all major pathologic categories, including neoplastic, traumatic, infectious, and inflammatory disorders. Using a validated AI-based annotation strategy \cite{Lyu2025-bt}, GPT-4.1-mini automatically assigned radiologic diagnoses to each study based on the radiology reports. These labels were not used for Vol-JEPA pretraining and were used solely to train and evaluate supervised diagnostic heads. Detailed dataset characteristics, including sequence types, image resolutions, and diagnosis distributions, are provided in Extended Data Fig. \ref{exfig:ex_data2}.

JEPAs are self-supervised learning methods that, given a context region within a data sample, predict a non-overlapping target region in a learned latent space. JEPA has achieved state-of-the-art performance on image and video data, but has not been applied to volumetric medical images \cite{Assran2023-iu, Assran2025-vf, Bardes2024-em}. We developed Vol-JEPA for volumetric neuroimages to capture invariant semantic information, such as radiologic diagnoses, and spatially equivariant local information, including lesion localization, patient orientation, and neuroanatomy. A critical component of JEPA training is the masking strategy used to define context and target regions. Because neuroimages exhibit intrinsic spatial structure determined by anatomy, we leveraged this property to design a neuroanatomy-informed masking strategy optimized for volumetric learning.

Given an MRI sequence or CT volume, the patient is first segmented as foreground, and the volume is divided into volume tokens (Fig. \ref{fig:fig_1_modeling}b). A random location within the patient volume is selected as the centroid of a masked block, and additional centroids are sampled until approximately 85\% of tokens are masked. The union of these regions forms the Vol-JEPA target, while the remaining tokens define the context (Extended Data Fig. \ref{exfig:ex_data1}). In one configuration, peripheral regions such as the face, scalp, and prefrontal cortex serve as context, and the model, a vision transformer \cite{Dosovitskiy2020-xs}, predicts deeper structures including the basal ganglia, brainstem, and occipital lobe. In the inverse configuration, the context consists of deep structures and the target is the periphery. Each minibatch contains an equal mix of both configurations to ensure balanced neuroanatomic learning. Successfully performing this task requires an understanding of anatomic orientation, neuroanatomy, radiographic features, and neurologic disease. Because Vol-JEPA learns through prediction rather than autoencoding or contrastive objectives, it does not require voxel-level augmentations, negative pairs, generative decoders, large batch sizes, or radiology reports. This enables efficient scaling to large, uncurated clinical datasets. A detailed schematic of Vol-JEPA training is shown in Extended Data Fig. \ref{exfig:ex_data3}.


\begin{figure*}[t!]
    \centering
    \includegraphics[width=\textwidth]{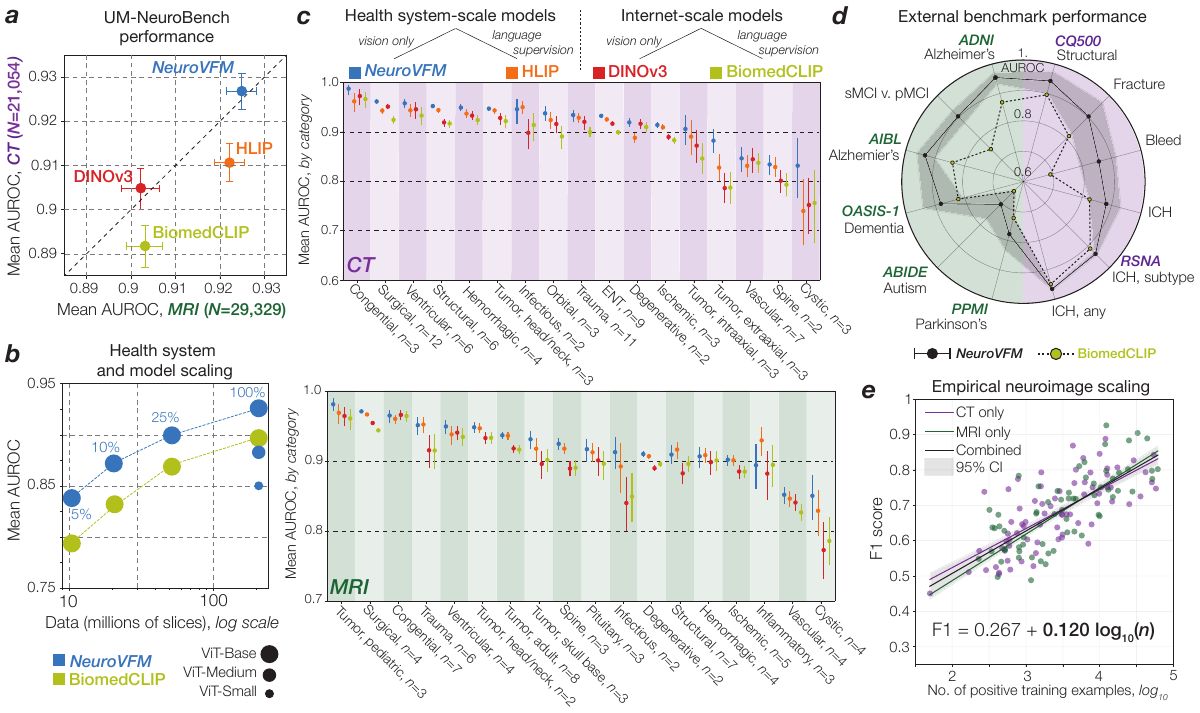} 
    \vspace{-15pt}
    \caption{\footnotesize \textbf{NeuroVFM results.}
    \textbf{a}, NeuroVFM performance over 82 CT and 74 MRI diagnostic tasks, and compared with both health system-scale (HLIP) and internet-scale (DINOv3, BiomedCLIP) models. NeuroVFM outperforms models trained with language supervision and those trained on public internet data. Results are mean \(\pm\) 95\% CI. \textbf{b}, NeuroVFM exhibits foundation model behavior, with performance scaling across data volume and model size. \textbf{c}, Performance across diagnostic ontologies, such as traumatic, congenital, ischemic lesion, etc. is shown for both CT and MRI. NeuroVFM consistently outperforms other baselines. \textbf{d}, NeuroVFM was tested on CT and MRI-based \emph{external} benchmarks, including Alzheimer's disease, Parkinson's disease, and autism classification, as well as intracranial hemorrhage detection. NeuroVFM outperformed internet-scale models by a wide margin. \textbf{e}, We discovered an empirical log-linear scaling relationship between the number of positive training examples and model performance. This relationship held across at least 4 orders of magnitude, imaging modalities, and models (Extended Data Fig. \ref{exfig:ex_data4} and \ref{exfig:ex_data5}).
    }
    \label{fig:fig_2_results}
    \vspace{-15pt}
\end{figure*}

\begin{figure*}[t]
    \centering
    \includegraphics[width=\textwidth]{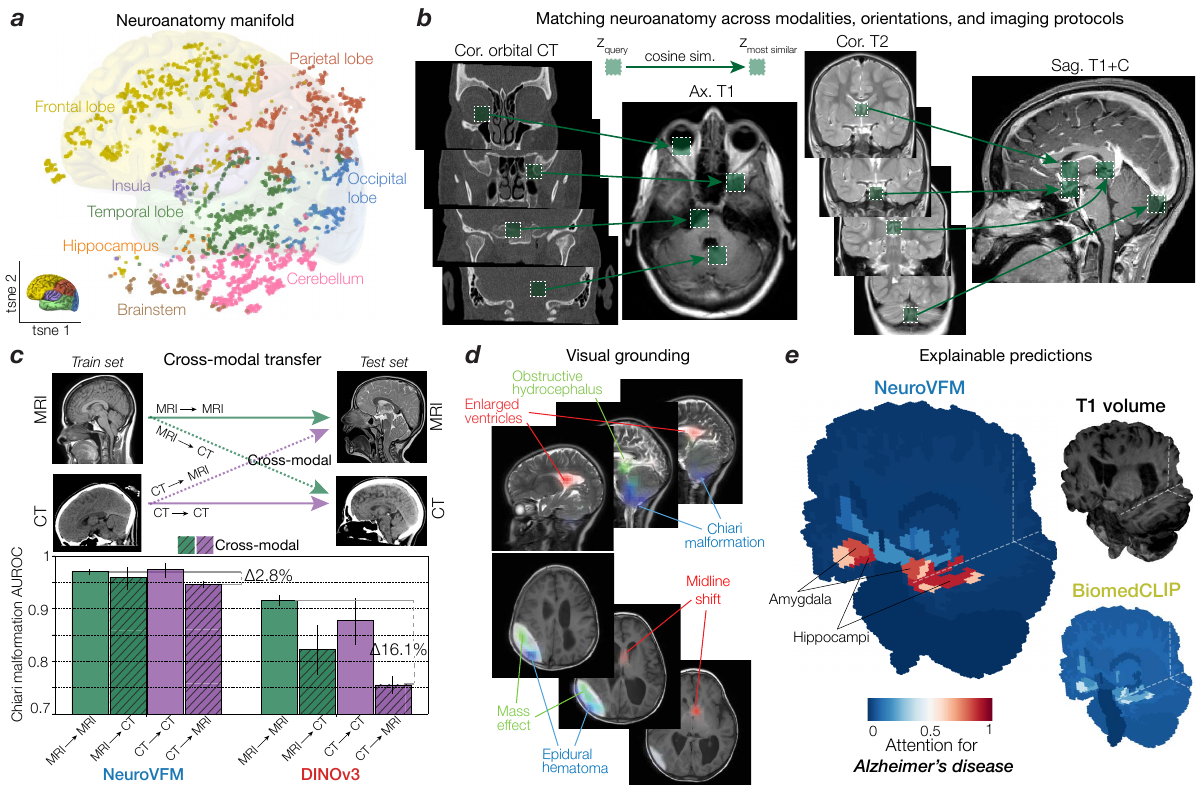} 
    \vspace{-15pt}
    \caption{\footnotesize \textbf{NeuroVFM understands neuroimaging.}
    \textbf{a}, A T1-weighted MRI sequence was segmented using the SynthSeg model \cite{Billot2023-gk}. Tokens were embedded into the learned NeuroVFM representation space and plotted using t-SNE. Each token is plotted (dots) and colored according to the segmentation map. NeuroVFM organized visual tokens into a latent neuroanatomic manifold that accurately encodes both spatial and semantic features. 
    \textbf{b}, NeuroVFM implicitly matches and registers anatomic regions from different modalities, patient orientations, and imaging protocols. 
    \textbf{c}, As a multimodal model with a shared representation space between CT and MRI, NeuroVFM understands pathology across these modalities. An MRI classifier for Chiari malformations, for example, transfers to CT, and vice versa. DINOv3 representations do not exhibit consistent cross-modal performance. Additional cross-modal transfer results are in Extended Data Fig. \ref{exfig:ex_data4} and \ref{exfig:ex_data5}. 
    \textbf{d}, NeuroVFM predictions are grounded, with diagnostic attention placed on pathologic regions. A patient with a history of obstructive hydrocephalus and Chiari malformation is above, and below is a patient with an epidural hematoma causing mass effect and midline shift. 
    \textbf{e}, NeuroVFM also produced explainable classifier predictions. A NeuroVFM-based Alzheimer's disease classifier attends to mesial temporal structures, regions most implicated in disease progression.
    }
    \label{fig:fig_3_scene}
    \vspace{-15pt}
\end{figure*}

\begin{figure*}[t!]
    \centering
    \includegraphics[width=\textwidth]{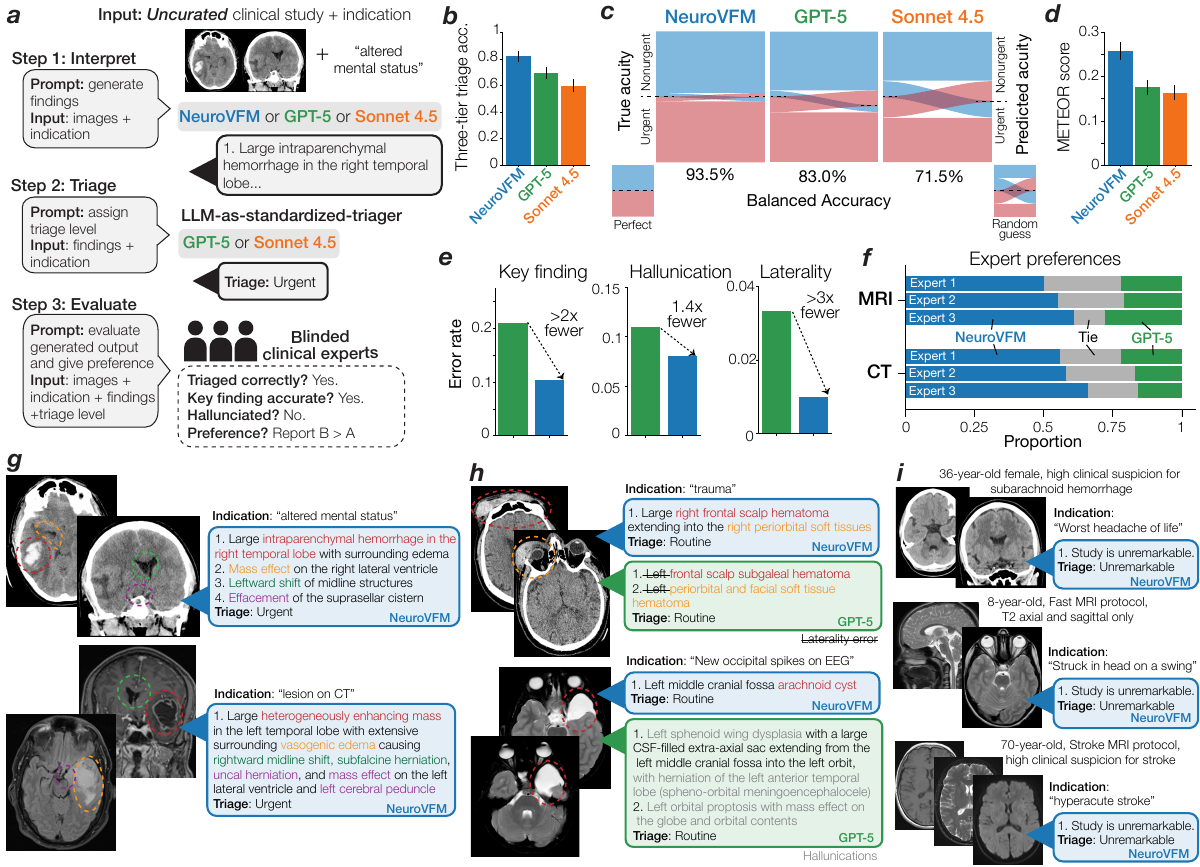} 
    \vspace{-15pt}
    \caption{\footnotesize \textbf{Triage report generation results.}
    \textbf{a}, Overview of the study design and workflow for report generation and triage. (Step 1) A generative multimodal model, such as NeuroVFM-LLaVa or GPT-5, is prompted to generate image key findings. (Step 2) An LLM, GPT-5 or Claude Sonnet 4.5, is then used as a standardized judge to triage the study based on the findings generated in Step 1. (Step 3) \emph{Blinded} clinical experts then evaluate whether the study was triaged correctly, the quality of the generated reports, and which report they prefer. NeuroVFM outperformed the other frontier models on overall triage performance (\textbf{b}), urgent findings generation (\textbf{c}), and all natural language processing metrics (\textbf{d} and Extended Data Fig. \ref{exfig:ex_data9}). \textbf{e}, Blinded clinical experts noted that NeuroVFM was more likely to generate correct key findings, less likely to hallucinate, and much less likely to make a laterality error compared to GPT-5. \textbf{f}, Clinical experts also preferred NeuroVFM reports more than 2:1 over GPT-5 reports. \textbf{g}-\textbf{i}, Illustrative examples of urgent, routine, and unremarkable head CTs and brain MRIs in the expert annotated testing set.
    }
    \label{fig:fig_4_vlm}
    \vspace{-15pt}
\end{figure*}

\subsection*{Testing NeuroVFM on diverse neuroimaging tasks}
To compare pretraining on health system data with pretraining on public data, we conducted a year-long, multicenter diagnostic study across our health system. All patients consecutively evaluated between June 1, 2023, and May 30, 2024, who underwent a CT or MRI of the head or neck were enrolled without exclusion, yielding a representative sample of real-world clinical imaging. This cohort comprised more than 21,000 head CTs and 29,000 brain MRIs (Supplementary Data Table \ref{exfig:supp_table1}).

NeuroVFM achieved an average AUROC of 92.7 $\pm$ 0.4\% across 82 CT diagnostic tasks and 92.5 $\pm$ 0.4\% across 74 MRI diagnostic tasks (Fig. \ref{fig:fig_2_results}a). We compared NeuroVFM with three state-of-the-art models: the health system–trained HLIP model \cite{Zhao2025-nv} and two internet-scale baselines, DINOv3 and BiomedCLIP. For each model, we trained an attentive pooling classifier on frozen embeddings from UM-NeuroImages. NeuroVFM outperformed all baselines for both CT and MRI diagnosis, demonstrating the value of scalable self-supervised learning and health system–scale pretraining. Its performance was comparable to or exceeded that of established diagnostic tools used by expert neuroradiologists such as the ASPECTS score for ischemic stroke \cite{Pexman2001-zt} and the T2-FLAIR mismatch sign for diffuse glioma \cite{Patel2017-pm} (Extended Data Fig. \ref{exfig:ex_data4} and \ref{exfig:ex_data5}).

We found that NeuroVFM exhibits the hallmark properties of foundation models, with performance scaling predictably across both data volume and model size (Fig. \ref{fig:fig_2_results}b) \cite{Kaplan2020-ia}. Empirically, we observed a non-saturating log-linear relationship between the number of positive diagnostic examples and performance metrics (Fig. \ref{fig:fig_2_results}e). NeuroVFM maintained stable performance across MRI manufacturers, magnetic field strengths, demographic subgroups, and medical centers (Extended Data Fig. \ref{exfig:ex_data6}). We further evaluated the model on multiple public neuroimaging benchmarks, including brain age estimation, dementia and autism classification, and intracranial hemorrhage detection. Across nearly all tasks, NeuroVFM outperformed internet-scale baselines by a wide margin (Fig. \ref{fig:fig_2_results}d, Extended Data Fig. \ref{exfig:ex_data7}).


\subsection*{Emergent neuroimage understanding}

Health system learning with Vol-JEPA gave rise to emergent abilities: clinically meaningful behaviors that were not explicitly supervised during training. Without segmentation or label supervision, NeuroVFM organized visual tokens into a latent neuroanatomic manifold that accurately encodes both spatial and semantic features of brain structures (Fig. \ref{fig:fig_3_scene}a). Token-level encoding allows the model to match anatomic regions and pathologies across modalities, orientations, and imaging protocols, effectively performing zero-shot semantic matching (Fig. \ref{fig:fig_3_scene}b). Similar to how token matching emerges in frontier models trained on natural images such as DINOv3, we observe analogous behavior in NeuroVFM. For example, the model recognizes that the pineal gland on coronal T2-weighted imaging corresponds to the same structure on sagittal T1-weighted post-contrast imaging. These results imply that NeuroVFM has learned an \emph{implicit brain atlas of human neuroanatomy}. This emergent capability extends to zero-shot diagnostic transfer between MRI and CT, a property not observed in DINOv3 embeddings (Fig. \ref{fig:fig_3_scene}c). A classifier trained on CTs to detect Chiari malformation, midline shift, or ventriculomegaly transfers to MRIs without additional training (Extended Data Fig. \ref{exfig:ex_data4}, \ref{exfig:ex_data5}). By learning invariant disease representations that decouple pathology from modality-specific features, NeuroVFM inherits robustness to variations in scanner hardware, protocols, and sites (Extended Data Fig. \ref{exfig:ex_data6}).

We define diagnostic grounding as the model’s ability to localize image regions that causally support a diagnostic prediction. To evaluate grounding, we used an attention-based multiple instance learning (AB-MIL) framework (see Methods) \cite{Ilse2018-ix}. NeuroVFM accurately localized pathologic regions and mapped them to corresponding radiologic diagnoses (Fig. \ref{fig:fig_3_scene}d). For example, NeuroVFM correctly attends to and assigns the diagnostic labels to a patient with an epidural hematoma causing mass effect and midline shift. This comprehensive visual understanding extends across diverse pathologies, as illustrated in Extended Data Fig. \ref{exfig:ex_data8}. NeuroVFM also produced explainable classifier predictions for complex diseases such as Alzheimer’s, attending to mesial temporal structures, including the hippocampus and amygdala, known to be implicated in disease progression (Fig. \ref{fig:fig_3_scene}e) \cite{Planche2022-ps}. These results parallel the emergent abilities seen in frontier foundation models \cite{Wei2022-iz}, suggesting that health-system-scale self-supervised training can yield qualitatively new capabilities in medical AI.

\subsection*{NeuroVFM outperforms frontier models for report generation}
Frontier models such as GPT-5 and Claude Sonnet 4.5 are generative vision-language systems trained at internet scale on a vast corpus of public medical images and text. We tested whether NeuroVFM, when paired with an open-source language model trained only with autoregressive next-word prediction, could outperform these frontier models on the task of generating radiology reports for clinical triage. Triage is a high-yield application for AI, benefiting patients, radiologists, and clinicians through increased efficiency and early prioritization of urgent findings \cite{Titano2018-mx, Wood2022-ph}. In this task, the model must generate itemized reports for \emph{real-world, uncurated, clinical, neuroimaging studies} that accurately describe all key pathologic findings to classify each study into one of three clinical categories: unremarkable, routine, or urgent (Fig. \ref{fig:fig_4_vlm}a).


The frozen NeuroVFM model was paired with the open-source language model Qwen3 in a standard LLaVA-1.5-style visual instruction tuning framework to evaluate the quality of NeuroVFM’s learned representations for report generation (Fig. \ref{fig:fig_4_vlm}) \cite{Yang2025-rb, liu2023improvedllava}. Our goal was not to optimize for report generation performance but to assess the feasibility of training a simple generative model directly from NeuroVFM embeddings. The vision-language model was trained using standard supervised fine-tuning on the UM-NeuroImages dataset to generate all key pathologic findings (Extended Data Fig. \ref{exfig:ex_data9}). We used frontier reasoning models GPT-5 and Claude Sonnet 4.5 as baselines, prompting each to generate key findings (Supplementary Data Fig. \ref{exfig:supp_data5}). A fixed prompt and language model (GPT-5 or Sonnet 4.5) were then used as standardized judges to assign triage levels based on the generated reports, providing a standardized and model-agnostic evaluation protocol (Supplementary Data Fig. \ref{exfig:supp_data6}). All models were tested on 300 expert-selected CT and MRI studies, balanced across modality and triage level.

NeuroVFM outperformed GPT-5 and Claude Sonnet 4.5 on both three-tier triage accuracy and detection of urgent findings (Fig. \ref{fig:fig_4_vlm}b, c). When compared with ground-truth radiology reports, NeuroVFM-generated findings achieved higher scores across all natural language evaluation metrics, including METEOR and ROUGE (Fig. \ref{fig:fig_4_vlm}d and Extended Data Fig. \ref{exfig:ex_data9}). In blinded expert evaluation, the key finding error rate of NeuroVFM-generated reports was roughly half that of GPT-5 (approximately 10\% versus 20\%), with substantially fewer hallucinations and laterality errors. An international cohort of three blinded clinical experts preferred NeuroVFM-generated reports more than two to one over GPT-5-generated reports with high inter-rater concordance (Fig. \ref{fig:fig_4_vlm}f; Fleiss’ \(\kappa = 0.718\)).

Figure \ref{fig:fig_4_vlm}g-i shows representative NeuroVFM-generated reports across urgent, routine, and unremarkable triage levels. NeuroVFM accurately identified large mass lesions, including intraparenchymal hematomas and infiltrative tumors, and recognized associated findings such as midline shift, effacement of the basal cisterns, and brain herniation to determine lesion severity. The model also distinguished intracranial from extracranial hematomas, such as scalp hematomas, an essential distinction for triaging traumatic brain injury. NeuroVFM correctly identified non-urgent findings such as arachnoid cysts, even when producing mild mass effect, and consistently triaged unremarkable CT and MRI studies across imaging protocols and age groups, including those with high-risk clinical indications such as “worst headache of life” or “acute stroke.” We observed that both NeuroVFM and GPT-5 appropriately used the clinical indication to improve triage performance, whereas Claude Sonnet 4.5 struggled when given images alone (Extended Data Fig. \ref{exfig:ex_data9}). Examples spanning the full range of pathologies encountered in routine clinical practice are shown in Extended Data Fig. \ref{exfig:ex_data10}.

%% file: latex/discussion.tex
\section*{Discussion}
Health systems are knowledge bases and data engines that capture the collective experience of public health and clinical medicine. Here, we demonstrate that directly experiencing the clinical world through health system learning achieves high-performance neuroimaging foundation models. These task-agnostic vision models, pretrained on large and diverse clinical datasets, provide robust and transferable representations for downstream tasks. By integrating domain-specific intelligence learned from private clinical data with general-purpose reasoning, health system learning and NeuroVFM provides a roadmap towards AI systems that interpret and act on clinical information with subspecialist depth, expert reliability, and safe deployment.


Frontier models benefit from the vastness of publicly available information; however, this breadth limits their depth in technical domains, such as medicine \cite{Randall2025-dn}. Internet data rarely reflects the complexity and diversity of real-world clinical imaging, where disease manifestations, acquisition protocols, and patient anatomy can vary widely. Health system learning represents a fundamentally different paradigm \cite{Jiang2023-zr, Shmatko2025-tt}. Rather than learning from descriptions of the clinical world or curated hand-selected patient examples, health system learners, like NeuroVFM, experience the world itself, modeling the raw signatures of diseases embedded in unfiltered clinical data. World models have gained attention as a strategy to address the limitations of sequence modeling and reinforcement learning \cite{Ha2018-dr, Hafner2025-gs, Garrido2024-mk}. NeuroVFM shows that intelligence can emerge from observing clinical practice itself.


Our long-term vision for NeuroVFM is to complement frontier models, not replace them in medicine. Agentic AI systems, including GPT-5 and Claude Sonnet 4.5, have integrated tool use to improve performance in specific domains, such as mathematics, science, and medicine \cite{OpenAI2024-ph, Guo2025-id, Lu2025-rb}. We foresee NeuroVFM and other health system models being integrated into agentic AI systems as external modules that provide a grounded understanding of clinical data. Frontier models will be able to reason over structured, clinically calibrated outputs from NeuroVFM, reducing failure modes and providing an auditable substrate for action. We believe expert-level performance across most cognitively complex domains will require the integration of domain-specific modules into general-purpose AI models.

Limitations and directions for future research include the need to integrate temporal and additional multimodal data streams, including pathology, genomics, and longitudinal clinical outcomes, to construct unified representations of disease progression. While the present study focuses on neuroimaging, the unified Vol-JEPA architecture is inherently extensible to other medical imaging modalities and body parts. Finally, while NeuroVFM exhibits emergent interpretability, translating these insights into actionable clinical interfaces will require new methods for human–AI collaboration, uncertainty quantification, and prospective evaluation in clinical workflows.

In summary, NeuroVFM demonstrates that generalist medical intelligence can be built from the health system itself. Learning directly from health system data yields representations that translate to diagnosis, triage, and report generation. We provide a scalable blueprint towards the development of medical foundation models that will be transformative in 21\textsuperscript{st} century healthcare.

%% file: latex/methods.tex
\clearpage
\section*{Methods}
\subsection*{Principles of Health System Learning} 
We view health system learning as the environment within which generalist medical foundation models can emerge \cite{Moor2023-av}. To clarify the concept and differentiate it from existing foundation model training methods, such as internet-scale or life science training, we elaborate on several principles that define health system learning.

1) \emph{Directly learning the clinical world}: Models should learn from data generated during routine patient care with no or minimal data processing or curation, experiencing the breadth and complexity of clinical medicine. This ensures robustness, realism, and coverage of clinically important edge cases.

2) \emph{Scalable training}: 
Health system datasets require scalable supervision. Self-supervised learning objectives, including predictive, contrastive, and autoencoding objectives, are the current preferred strategy. However, other methods, such as semi-supervision or energy-based objectives, are candidates for scalable health system learning.

3) \emph{Clinical grounded representations}: 
Medical foundation models trained through health system learning should internalize anatomy, pathology, and physiologic variation that supports accurate diagnosis, cross-modal reasoning, and interpretable grounding, thereby reflecting the epistemic structure of clinical expertise.

4) \emph{Multimodal medical learning}: Health system data are multimodal representations of a single underlying object: the patient. Any single modality is a partial and incomplete view of the patient. Multimodal models can achieve more complete representations and better disentangle explanatory features across modalities

5) \emph{Integration with agentic and multimodal AI systems}: 
Health system learning is inherently modular. The resulting domain-specific foundation models can serve as expert visual or multimodal modules within larger agentic systems. This complements general-purpose models, enabling clinically safe, domain-aligned reasoning at scale.


\subsection*{UM-NeuroImages Dataset} 
Radiology studies at large academic medical centers are stored in picture archiving and communication systems (PACS). We queried the University of Michigan PACS via a SQL-based interface (Sectra Data Warehouse) for neuroimaging studies that satisfy: 1) an acquisition date prior to June 1, 2023; 2) an examination of body part(s) including the head, brain, orbits, face, or neck; and 3) a modality of MRI or CT. The specific SQL query parameters are in Supplementary Data Figure \ref{exfig:supp_data1}.

The initial query yielded 645,989 unique studies. After removing non-image/corrupted entries and embedded non-DICOM files, the final cohort comprised 566,915 studies (275,981 MRI studies and 290,934 CT studies). Series counts were 3,647,950 MRIs and 1,591,629 CTs, for a total of 5,239,579. Descriptive characteristics are in Extended Data Fig. \ref{exfig:ex_data2}a-d. For 444,188 studies with paired radiology reports (MRI, 218,882; CT, 224,306), we used GPT-4.1-mini to extract diagnoses and summarized findings, following prior report-parsing work \cite{Adams2023-ob, Titano2018-mx, Chien2024-ls, Ranjit2023-hk}. We randomly held out approximately 1,000 MRI and 1,000 CT studies as a validation set for model selection. The test cohort is prospective and temporally held-out, including studies from June 1, 2023, through May 31, 2024.

\subsection*{Series Preprocessing}
To ingest the full breadth of health system data without manual curation, we applied an automated pipeline to standardize intensities and dimensions. Series were resampled to \(1\times1\times4\,\mathrm{mm}\) (\(4\,\mathrm{mm}\) along the acquisition axis from DICOM) and saved as 8-bit. MRI intensities were clipped to the 0.5-99.5th percentile before quantization. CT volumes yielded three windows: brain (width=80, level=40; 8-bit), subdural/blood (width=200, level=80; 4-bit), and bone (width=2800, level=600; 4-bit). 4-bit windows reduced storage while preserving contrast for key pathologies. Background masks were derived via Otsu and Hounsfield thresholding for MRI and CT, respectively. During training, volumes were cast to float, scaled to [0,1], and mean-normalized using statistics derived from the training set. Extended Data Fig. \ref{exfig:ex_data1}b summarizes this pipeline.

\subsection*{NeuroVFM Training with Vol-JEPA}
The core training objective for NeuroVFM was large-scale self-supervision, applied at the individual volume level, across the entire UM-NeuroImages dataset. NeuroVFM employs a joint-embedding predictive architecture (JEPA) to optimize a masked modeling objective directly in the representation space. First, input volumes are tokenized into non-overlapping 3D patches of 4×16×16 voxels. Each 3D volume is then partitioned into a small, visible context region, $x$, and larger, masked target region, $y$. The tokenized context region is processed by a trainable student encoder, $E_\theta$. The resulting context representations $E_\theta(x)$, along with positional information of the masked target patches, are then fed to a predictor module, $P_\phi$, which generates predictions for the representations of these target patches. The complete set of tokens from the input volume (representing both context and target regions) is processed by a teacher encoder, whose weights $E_{\bar{\theta}}$ are updated as an exponential moving average (EMA) of the student encoder's weights after each step and are not directly optimized by gradient descent. The training objective minimizes the difference between the predicted representations of the target patches and those generated by the teacher network for the same target patches. This is formulated as:

$$\text{minimize}_{\theta,\phi,\Delta_y} \hspace{1mm} \lVert P_\phi(\Delta_y, E_\theta(x)) - \text{sg}(E_{\bar{\theta}}(y)) \rVert_{\text{smooth L1}},  \hspace{3mm} \text{\cite{Assran2023-iu}}$$ 

where $\Delta_y$ is a set of learnable tokens that represents the masked target patches and $\text{sg}(\cdot)$ denotes a stop-gradient operation. This representation-level objective is computationally efficient and encourages the learning of semantic features without requiring explicit voxel-level augmentations or decoding.

Volumetric-JEPA (Vol-JEPA) extends the principles of I-JEPA \cite{Assran2023-iu} and V-JEPA \cite{Bardes2024-em} for self-supervision to volumetric neuroimages, predicting representations of masked 3D target patches based on visible 3D context patches. This encourages the model to learn the shared anatomy of the brain, head, and neck. The masking strategy was foreground-focused: context and target patches were sampled exclusively from the pre-computed head mask. Masking was performed in two ways: (1) multiple large crops are sampled, with their union forming the masked target, and (2) a small crop is sampled as context, with the complement being the target. A random subset of context patches is then dropped, serving as additional target patches to predict. Unlike approaches that truncate sequences within a mini-batch to a uniform length, our implementation leverages FlashAttention-2 \cite{Dao2023-sb} to directly encode variable-length sequences of context and target patches derived from input volumes. To ensure robustness to varying patient orientations, we applied random axis permutations and flips to each volume, using the same transform for both student and teacher inputs. A hyperparameter search indicated optimal context region sampling ratios (from the total patches within a given crop) of 25\% for MR volumes and 20\% for CT volumes, with a patch dropout rate of 20\%. An overview of this masking strategy can be found in Extended Data Fig. \ref{exfig:ex_data3}a.

During training, volumes were truncated along each axis to a maximum of 20 patches to bound the token count per crop. At inference, the same encoder was applied to the entire volume without truncation, leveraging its ability to handle variable-length token sequences. For CT scans, which are typically viewed with multiple windowing presets (e.g., brain, subdural, bone), we implemented a weighted sampling strategy during training. For pretraining, one of these three CT window settings was randomly selected and applied with probabilities of 0.7 (brain), 0.15 (subdural), and 0.15 (bone), and all windows were used during inference. Full training details are in Supplementary Data Table \ref{exfig:supp_table6}.

\subsection*{Evaluating diagnostic grounding with multiple instance learning}
Most visual grounding evaluation requires an object detection module to output a bounding box around an image region given a label or text prompt. This is not feasible for the current study because there are no object detection datasets or models sufficiently powerful to evaluate neuroimage grounding on the scale and complexity of UM-NeuroImages. Because most neurologic pathologies are spatially small relative to the full study, study-level labels provide only weak supervision. We leveraged an attention-based multiple instance learning (AB-MIL) framework to assess neuroimage grounding \cite{Ilse2018-ix}. AB-MIL is known to assign high attention to diagnostic regions in medical images \cite{Chen2024-up}. Unfortunately, the standard AB-MIL framework, with its `aggregate-then-classify' design, cannot resolve a critical grounding ambiguity: it is unable to disentangle a patch's importance, \textit{where} the model looks, from its directional contribution, \textit{why} it is considered positive or negative evidence.

We address this gap with a pooling operation that reverses the standard order to a `classify-then-aggregate' model. For a task with $K$ classes and a bag of $N$ instances, we first compute classification logits for each instance $i$ using a multilayer perceptron (MLP), $\psi_{p}$. A separate attention MLP, $\psi_{m}$, generates $K$ class-specific attention scores per instance. For each class, these scores are normalized with a softmax across instances to obtain class-specific attention weights. The final bag-level logits, $p(x)$, are the sum of the element-wise product of the per-instance logits and their corresponding attention weights:

$$p(x) = \sum_{i=1}^{N} \alpha_{i} \circ \psi_{p}(f(x_{i}))$$

Here, $f(x_i)$ is the frozen feature vector for instance $i$, $\alpha_i$ is the vector of class-specific attention weights, and $\circ$ denotes the Hadamard product. This formulation yields interpretable, label-specific attention maps that reflect both the importance and directional contribution of each region to the diagnostic decision, providing a scalable means of evaluating grounding without region-level annotations.

\subsection*{Comparison to Foundation Model Baselines}

We evaluated NeuroVFM against two families of baselines: (1) Internet-scale pretrained encoders, and (2) methods directly trained on UM-NeuroImages. All backbones were frozen and evaluated with the same study-level attentive probe and data splits. Full architecture details, training hyperparameters, and evaluation configurations can be found in Supplementary Data Table \ref{exfig:supp_table6} and our GitHub repository. A schematic of the study-level pooling strategies can be found in Extended Data Fig. \ref{exfig:ex_data1}e.

\subsubsection*{Internet-scale pretrained encoders}

We chose two baselines representative of the dominant Internet-scale paradigms:
\begin{itemize}
    \item Natural image self-supervision, \textbf{DINOv3} \cite{Simeoni2025-xo}: ViT-B/16 pretrained on 1.7 billion natural images (HuggingFace, `facebook/dinov3-vitb16-pretrain-lvd1689m`)
    \item Medical image-text alignment, \textbf{BiomedCLIP} \cite{zhang2023biomedclip}: ViT-B/16 pretrained on 15 million image-caption pairs scraped from PubMed (HuggingFace, `microsoft/BiomedCLIP-PubMedBERT\_256-vit\_base\_patch16\_224`)
\end{itemize}

For 2D encoders, 3D volumes were processed slice-wise according to the model's respective preprocessing pipeline (e.g., DINOv3 necessitates resizing to $224\times224$ and normalizing with ImageNet mean/std), with slice-level features aggregated to produce study-level predictions.

\subsubsection*{Methods trained on UM-NeuroImages}

To provide a controlled comparison of representation learning methods, we trained key architectures on our UM-NeuroImages dataset. For comparison to vision-only self-supervision strategies, we benchmarked against a 3D \textbf{Masked Autoencoder} (VideoMAE) \cite{Tong2022-nt} with an 85\% foreground-aware masking ratio, and \textbf{DINOv2} model \cite{Oquab2023-co} trained on 2D neuroimaging slices derived from UM-NeuroImages volumes. The DINOv2 features were aggregated in a similar fashion to the pretrained 2D vision encoders. For vision-language alignment comparison, we trained \textbf{HLIP} \cite{Zhao2025-nv}, a state-of-the-art 3D medical vision-language model, on our 444,188 study-report pairs and evaluated its study image encoder. HLIP training minibatches were balanced across MRI and CT studies, and the CLIP objective was computed within each modality to prevent trivial cross-modality discrimination.

\subsection*{Diagnostic Evaluation of NeuroVFM}

The UM-NeuroImages dataset contains two decades of clinical neuroimaging data from a large academic health system, spanning a broad spectrum of neurological presentations. To evaluate \emph{imaging-only} diagnostic ability, we evaluated NeuroVFM on a prospective, temporally held-out test cohort beginning immediately after the retrospective window. Labels were derived from radiology reports using a LLM extraction pipeline, with a subset manually verified by clinical experts. We defined a clinically organized ontology of 74 MRI and 82 CT diagnoses (Extended Data Fig. \ref{exfig:ex_data2}h). Encoders were frozen, and modality-specific attentive probes were trained for multi-label prediction using class-weighted binary cross-entropy. Hyperparameters and class-specific thresholds were selected on the validation set. We report per-class AUROC, balanced accuracy, sensitivity, and specificity on the test cohort, with 95\% CIs from study-level bootstrap resampling (10,000 replicates). Full results are provided in Supplementary Data Table \ref{exfig:supp_table2} and \ref{exfig:supp_table3}.

\subsubsection*{Data, model, and modality scaling}

We studied performance along data scale, model capacity, and modality with resource-normalized protocols. For modality, we held the optimization budget fixed at \(\sim\)510K training steps and compared (i) a single multimodal model vs. (ii) two unimodal models (one MRI, one CT) trained separately. For data and model scaling, we varied (a) the fraction of UM-NeuroImages (5\%, 10\%, 25\%), and (b) the encoder size (ViT-Small, ViT-Medium, ViT-Base). Larger backbones (e.g., ViT-Large) were left to future work due to computationally prohibitive hyperparameter search.

\subsubsection*{Label efficiency}

We quantified how per–task performance scales with the amount of supervision. For each diagnostic class $c$ and modality $m \in \{\mathrm{CT},\mathrm{MRI}\}$, we calibrated model probabilities on the validation set using Platt scaling and computed F1 on the held-out test set. We excluded classes with fewer than 30 training positives ($n_{\text{pos}}$) and 10 testing positives. We fit a simple linear model of F1 vs. $\log_{10}(n_{\text{pos}})$ via OLS with HC3 SEs. Modality fixed effects and interactions assessed CT vs. MRI differences. Encoder comparisons (NeuroVFM vs. HLIP, DINOv3, and BiomedCLIP) used ANCOVA to test slope equality. If interaction terms were non-significant, a shared slope was used, otherwise encoder-specific slopes were retained. Label-equivalence was reported as the fold-increase in positives a baseline requires to match NeuroVFM at a fixed F1, with 95\% CIs from 10{,}000 bootstrap replicates over classes.

\subsection*{NeuroVFM on External Public Benchmarks}

To assess the out-of-distribution generalization, we evaluated frozen NeuroVFM performance on eight public neuroimaging benchmarks. For all tasks, we trained an attentive probe without updating the encoder. Within each dataset, we held out 20\% of subjects as a stratified test set. On the remaining 80\%, we performed 8-fold iterative stratified cross-validation to select probe hyperparameters. The 8 probes trained on the cross-validation folds were then used to generate logits on the held-out test set, which were averaged per study to form the final ensemble. All reported metrics include 95\% subject-level percentile CIs (10,000 bootstrap replicates). This approach rigorously tests the quality of the learned representations and, through the probe's attention weights, allows for the identification of class-discriminative tokens for each task.

\subsubsection*{External MRI benchmarks}

We evaluated NeuroVFM on six public MRI datasets, spanning a range of neurological and psychiatric conditions. On the multi-site \textbf{OpenBHB} dataset \cite{Dufumier2022-am}, we benchmarked brain age regression to test for fine-grained anatomical representation. For dementia-related tasks, we utilized the ADNI dataset \cite{petersen2010alzheimer} to perform CN vs. Alzheimer's Disease (AD) classification and to distinguish progressive from stable Mild Cognitive Impairment (sMCI vs. pMCI), with a 20\% stratified test set held out within ADNI. We then evaluated the ADNI-trained AD classifier externally on the OASIS-1 \cite{Marcus2010-zk} and AIBL \cite{Fowler2021-qb} datasets, applying the frozen probe without further fine-tuning to distinguish cognitively normal individuals (CDR = 0) from those with dementia (CDR $\geq$ 1). We further evaluated diagnostic classification on several consortium datasets: differentiating individuals with Autism Spectrum Disorder from typically developing controls on the \textbf{ABIDE} dataset \cite{Di-Martino2014-mc} and Parkinson's Disease from healthy controls on the \textbf{PPMI} dataset \cite{Marek2018-ow}.

\subsubsection*{External CT benchmarks}

Performance on detecting critical neuroradiological findings was evaluated on two public non-contrast head CT cohorts, selected to test generalization on both a large-scale challenge dataset and a smaller, deeply-annotated clinical cohort. The large-scale 2019 \textbf{RSNA-ICH} Challenge dataset \cite{Flanders2020-pk} was used to benchmark multi-label classification of intracranial hemorrhage and its five subtypes (epidural, intraparenchymal, intraventricular, subarachnoid, and subdural). The high-quality, expert-annotated \textbf{CQ500} dataset \cite{Chilamkurthy2018-ow} was used for a more extensive evaluation across 14 diagnostic labels, including detailed hemorrhage characterization (subtypes, laterality, chronicity), skull fracture detection, and signs of structural abnormality (e.g., mass effect and midline shift).

\subsection*{Anatomical Probing of NeuroVFM}
To test whether NeuroVFM's frozen representations align with neuroanatomy, we paired unsupervised feature analyses with reference neuroanatomical segmentations, which were derived from SynthSeg \cite{Billot2023-gk}. First, we examined the low-dimensional organization of patch embeddings by running t-SNE on features from representative neuroimages and annotating each point by the majority SynthSeg region (frontal, parietal, temporal, occipital, insula, hippocampus, brainstem, and cerebellum) within its patch. We additionally tested whether NeuroVFM learns an approximate shared anatomical coordinate across subjects, modalities, sequences, and orientations via a \textit{patch-matching} task. For each query patch from a neuroimage $A$, we searched over all patches in a different neuroimage $B$ (different subject and acquisition) by L2-normalizing features and identifying the nearest neighbor by cosine similarity. An imaging expert defined a set of canonical neuroanatomical regions (e.g., sella, pineal region, torcula), and we counted a match when both patches lay within the same expert-defined region in their respective scans, without any explicit spatial alignment or supervision. This effectively tests for \emph{zero-shot} anatomical registration capabilities emergent from the self-supervised objective. Finally, we evaluated whether coarse neuroanatomy emerges without supervision by clustering dense, overlapping patch embeddings computed via sliding-window inference. A k-means model ($k=3$) fit on embeddings from the initial window defined canonical clusters that were propagated to all windows and combined into a high-resolution map by voxel-wise majority vote. To identify the \textit{brain parenchyma} cluster, we computed the IoU between each cluster mask and the SynthSeg brain mask, ultimately selecting the cluster with the highest IoU. Performance was summarized as the mean highest IoU across 30 neuroimages (10 each of axial, sagittal, and coronal natively) from the PPMI dataset, chosen as an external dataset with the full head and neck tissues included.

\subsection*{Vision-Instruction Tuning for Radiology Report Generation}

To evaluate NeuroVFM's potential as a visual backbone for multimodal applications, we adapted a LLaVA-1.5-style visual instruction-tuning framework \cite{liu2023improvedllava}. This experiment was designed not to optimize report generation performance, but rather to assess the feasibility of coupling NeuroVFM's learned representations with a large language model (LLM) with minimal architectural modifications.

\subsubsection*{NeuroVFM-LLaVA architecture}

NeuroVFM-LLaVA comprises three components: (1) the frozen NeuroVFM visual encoder, (2) an open-source LLM (Qwen3-14B) \cite{Yang2025-rb}, and (3) a connector module to bridge them.

Standard LLaVA connectors use a 2-layer MLP to project visual features into the LLM's word embedding space. This approach is insufficient for the high dimensionality of multi-sequence neuroimaging, where a single study may comprise a large and variable number of visual tokens (\(\geq\)20K). To address this, our connector module first employs a perceiver that operates sequence-wise (e.g., on T1, T2, FLAIR independently) \cite{Alayrac2022-jg, Jaegle2021-vb}. The perceiver aggregates the variable-length token sequence from each scan into a fixed-length representation of 64 latents. These fixed-length latents (total latents = 64 × $num\_sequences$) are then concatenated and passed to a 2-layer MLP projector.

\subsubsection*{Training Dataset and Curation}

The training dataset was derived from 444,188 unique neuroimaging studies. To create the text pairs, original radiology reports were summarized using GPT-4.1-mini to extract a concise list of key radiological findings.

Acknowledging that naive training on imbalanced medical datasets can degrade performance, we curated the training set via data resampling. Using the previously extracted diagnostic labels from each report, we performed weighted random sampling with replacement, assigning each study a weight inversely proportional to the prevalence of its rarest associated label. This process, which aims to balance the representation of less common pathologies, resulted in the final training dataset of $\sim$270K unique image-text pairs (444,188 total including duplicates).

\subsubsection*{Training Strategy}

Following the LLaVA methodology, training proceeded in two stages.

\begin{enumerate}
    \item Stage 1 (Connector Pre-training): Only the connector module (perceiver and MLP) weights are updated. The model was trained to map NeuroVFM image features to the corresponding reference key findings, which were formatted as a single, concatenated string.
    \item Stage 2 (Full Fine-tuning): Both the connector and the LLM weights are updated. In this stage, the model was trained on an instruction-following task. The input prompt was fixed to: "Generate a concise report of the key positive findings for this study." The target output was the same set of findings, but formatted as a structured JSON list ordered by clinical relevance.
\end{enumerate} 

Further details on training hyperparameters are provided in Supplementary Data Table \ref{exfig:supp_table7}.

\subsection*{Evaluation of report generation and clinical triage}

We evaluated NeuroVFM-LLaVA against two proprietary multimodal frontier models: GPT-5 (\textit{gpt-5-2025-08-07}, "reasoning" medium, "verbosity" low) and Claude Sonnet 4.5 (\textit{claude-sonnet-4-5-20250929}, "thinking" enabled). This evaluation assessed two criteria: (1) the factual accuracy of generated findings and (2) the clinical utility of these findings for a downstream triage task. To support this evaluation, we established Business Associate Agreements (BAAs) with vendors offering commercially available HIPAA-compliant access to frontier models, specifically OpenAI and Anthropic. This enabled the secure exchange of protected health information (PHI) while preserving patient privacy.

\subsubsection*{Benchmark Model Adaptation (3D-to-2D Conversion)}

A primary challenge in benchmarking against proprietary models is that they accept only 2D images and have limitations on total input image tokens. To create a fair comparison with our 3D-native model, we developed a systematic 3D-to-2D conversion pipeline.

\begin{enumerate}
    \item All 3D volumes (preprocessed with the NeuroVFM pipeline) were converted into $224\times224$ pixel 2D slices by resampling along the axis of acquisition and keeping every other slice.
    \item Non-diagnostic and derived sequences (e.g., scout images, phase/magnitude maps) were excluded.
    \item Due to API token limits, inputs were constrained to 15 sequences per study for GPT-5 ($\sim$360 slices) and 4 for Claude Sonnet 4.5 ($\sim$96 slices).
    \item If a study exceeded this limit, sequences were deterministically prioritized based on clinical relevance (i.e., for MRI: post-contrast T1, DWI/ADC, FLAIR, SWI, T2, non-contrast T1; for CT: at least one brain, blood, and bone window).
    \item All selected slices were converted to float and normalized between [0,1], and slices with >90\% black pixels were dropped.
\end{enumerate}

\subsubsection*{Evaluation of Generated Reports}

All generative models were evaluated on "UM-NeuroImages-Triage", a new, manually-curated test set of 600 studies (300 validation, 300 holdout) derived from our prospective test set. This set was hand-selected by neuroimaging experts to be balanced across three acuity classes ("Unremarkable", "Routine", "Urgent") and two modalities (MRI, CT).

We assessed report quality through automated metrics, BLEU-2 \cite{Papineni2001-om}, ROUGE-L \cite{Lin2004-qy}, and METEOR \cite{Banerjee2005-tm}, as well as with human evaluation. To ensure effective blinding, all model outputs were standardized to a uniform text format. For NeuroVFM-LLaVA and GPT-5, a neuroimaging expert (T.H.) performed a structured review of each study to quantify (1) capture of key findings, (2) clinically significant hallucinations, and (3) laterality errors. To evaluate preferences, three neuroimaging experts from the United States and Europe (R.J., A.K.M., T.H.) then performed a blinded, randomized pairwise review comparing generated findings from NeuroVFM-LLaVA and GPT-5 against the ground truth clinical report. For each case, model outputs were randomly labeled "Report A" and "Report B". Evaluators selected their preferred report ("Report A", "Report B", or "Both") based on overall clinical utility for triaging. Because Claude Sonnet 4.5 exhibited substantially lower triage accuracy on UM-NeuroImages-Triage, we restricted expert review and preference testing to NeuroVFM-LLaVA and the strongest baseline (GPT-5).

To quantify the clinical utility of the generated findings, we employed an "LLM-as-a-judge" pipeline. In each analysis, we designated a frontier model (GPT-5 or Claude Sonnet 4.5) as a separate "triager LLM", instructed with a strict, pre-defined set of criteria (Supplementary Data Fig. \ref{exfig:supp_data6}). We first confirmed that each triager achieved high accuracy when classifying acuity from the ground-truth radiology report alone. The triager LLM was then prompted to classify each study as "Unremarkable", "Routine", or "Urgent" based solely on the generated text findings from each model (NeuroVFM-LLaVA, GPT-5, and Claude Sonnet 4.5) and the clinical indication. Using GPT-5 vs. Claude Sonnet 4.5 as the triager yielded comparable results, indicating that the protocol is model-agnostic and not biased toward any single frontier model. This procedure converts report generation into a 3-class classification problem and provides a quantitative measure of clinical utility.

\subsection*{Computational hardware and software}
All data used to train NeuroVFM were acquired over more than two decades of routine clinical care at Michigan Medicine. Studies were identified in the Michigan Medicine PACS via the SDW. The aggregate size of raw NIfTI volumes is \(\sim\)150\,TB, stored on the ARC DataDen object-storage service. Post-processing (removal of non-image/corrupt entries and embedded metadata, background removal, and quantization) reduced the footprint by over an order of magnitude to \(\sim\)9\,TB. Preprocessed volumes were written as large NumPy memory-mapped shards (10 shards for the training set; 1 each for the validation and test set) with JSON index manifests to enable zero-copy random access during training.

All experiments were executed on the University of Michigan Advanced Research Computing (ARC) Armis2 cluster using SLURM. Typical jobs used nodes with 8 NVIDIA L40S GPUs (48\,GB each), 64 Intel Xeon Platinum 8358 CPU cores, and 503\,GB RAM. Vol-JEPA pretraining used PyTorch DistributedDataParallel (DDP), whereas LLM fine-tuning used PyTorch Fully Sharded Data Parallel (FSDP). A full Vol-JEPA pretraining run required \(<\)1{,}000 GPU-hours in aggregate (batch size of 768, \(\sim\)510K steps), making the approach feasible on university-hosted compute clusters. Training used mixed precision (AMP) with fixed random seeds.

Models were implemented in Python 3.10.14 with PyTorch 2.5.0 (CUDA 12.4). Vol-JEPA pretraining and diagnostic probing used PyTorch Lightning 2.5.0.post0. All data handling and computation occurred on HIPAA-compliant, access-controlled ARC infrastructure. Model weights are available under the CC-BY-NC-SA-4.0 license, enabling the broad dissemination of learned clinical representations without the risk of PHI leakage.

\subsection*{Use of large language models}

Large language models were used to assist with code prototyping and manuscript editing. All analyses, text and code were reviewed and verified by the authors, who take full responsibility for the work.

\section*{Ethics and inclusion statement}
Our research was approved by the University of Michigan Institutional Review Board (HUM00229133). All MRI and CT data were acquired under secondary data usage. The methods were carried out in accordance with the IRB’s guidelines, regulations, and policies. All human subjects that met inclusion criteria as stated above were included in the study. 

\section*{Data availability}
Institutional Review Board approval was obtained from University of Michigan for MRI data collection. Restrictions apply to the availability of raw patient MRI and CT imaging data, which were used with institutional permission through IRB approval for the current study and thus are not publicly available. All data sharing between medical centers is regulated through data use agreements with the study authors. A similar data-sharing protocol may be established for interested investigators. Please contact the corresponding author (T.H.) for any requests for data sharing. All requests will be evaluated based on institutional and departmental policies to determine whether the data requested is subject to intellectual property or patient privacy obligations. Data can only be shared for non-commercial academic and investigational purposes.

\section*{Code availability}
All code was implemented in Python (version 3.10.14) using PyTorch (2.5.0) compiled with CUDA 12.4 as the primary machine learning framework. The following packages were used for data preprocessing, model training and evaluation: pydicom (2.4.4), nibabel (5.3.2), SimpleITK (2.4.0), torchvision (0.20.0), pandas (2.2.3), NumPy (2.1.2), PyTorch Lightning (2.5.0.post0), flash-attn (2.6.3), matplotlib (3.10.7), scipy (1.15.2), and scikit-learn (1.6.1). The following packages were used to load baselines: open-clip (2.23.0) and transformers (4.56.0). All code and scripts to reproduce the experiments of this paper are available on GitHub at \url{https://github.com/MLNeurosurg/neurovfm} under an MIT license. 

\section*{Acknowledgments}
We would like to thank Karen Eddy, Muhammad Bhalli, Gary Laderach, and Brock Palen for providing technical support; David Hanauer for support with the University of Michigan Electronic Medical Record Search Engine (EMERSE); and Anthony Rosenzweig for scientific guidance. This work was supported by the following National Institute of Health (NIH) funding sources: K12NS080223 (T.C.H.), T32GM007863 (A.K., A.R.), and T32GM141746 (A.K.). This work was supported by the Chan Zuckerberg Foundation (CZI) Advancing Imaging Through Collaborative Project grant (T.H.), Cook Family Brain Tumour Research Fund (T.H.), Mark Trauner Brain Research Fund, Zenkel Family Foundation (T.H.), Ian’s Friends Foundation (T.H.), UM Precision Health Investigators Awards grant program (T.H.), Translational AI Award from the UM Department of Neurosurgery (T.H.), UM Stanley and Judith Frankel Institute for Heart and Brain Health Innovative Multidisciplinary Research Pilot Award (T.H.) and UM Research Scouts program (T.H.). This research was also supported, in part, through computational resources and services provided by Advanced Research Computing, a division of Information and Technology Services at the University of Michigan.

\section*{Author contributions}
A.K., A.R., and T.H. contributed to the conceptualization, study design, and analysis of results. A.K., A.R., C.Z., Y.L., S.H., S.B., R.H., C.J., A.C., and T.H. contributed to the experimentation, acquisition, analysis and interpretation of data. A.K., A.R., and T.H. contributed to generating the figures and tables for the manuscript. R.J., A.K.M., and T.H. provided neuroimaging expertise to evaluate generations. All authors were involved in the editing, analysis and review of all data and manuscript versions.

%% file: latex/extended_data.tex
\clearpage
\appendix

\section{Extended Data Figures}
\begin{enumerate}
    \item Extended NeuroVFM workflow
    \item Overview of UM-NeuroImages
    \item Learning with Vol-JEPA
    \item Extended CT diagnostic results
    \item Extended MRI diagnostic results
    \item Subgroup and health system analysis
    \item Open benchmark performance
    \item Neuroimage scene understanding and grounded diagnoses
    \item Generative evaluation
    \item Generated triage reports
\end{enumerate}

\section{Supplementary Data}

\subsection{Supplementary Data Figures}
\begin{enumerate}
    \item SQL query for UM-NeuroImages curation
    \item MRI diagnoses extraction prompt
    \item CT diagnoses extraction prompt
    \item Prompts for structured findings extraction
    \item Frontier model prompt for study findings
    \item Triage prompt given findings and indication
\end{enumerate}

\subsection{Supplementary Data Tables}
\begin{enumerate}
    \item Descriptive characteristics of the UM-NeuroImages prospective test set
    \item Full task-level performance on the UM-NeuroImages CT test set
    \item Full task-level performance on the UM-NeuroImages MRI test set
    \item Performance of NeuroVFM on external neuroimaging benchmarks
    \item Generated reports and evaluations for the UM-NeuroImages-Triage test set
    \item Hyperparameters for NeuroVFM pretraining and UM-NeuroImages study-level attentive probing
    \item Hyperparameters for NeuroVFM-LLaVA training and inference
\end{enumerate}

\clearpage
\renewcommand{\figurename}{Extended Data Figure}
\setcounter{figure}{0}

\begin{figure*}[p!]
    \centering\includegraphics[scale=0.78]{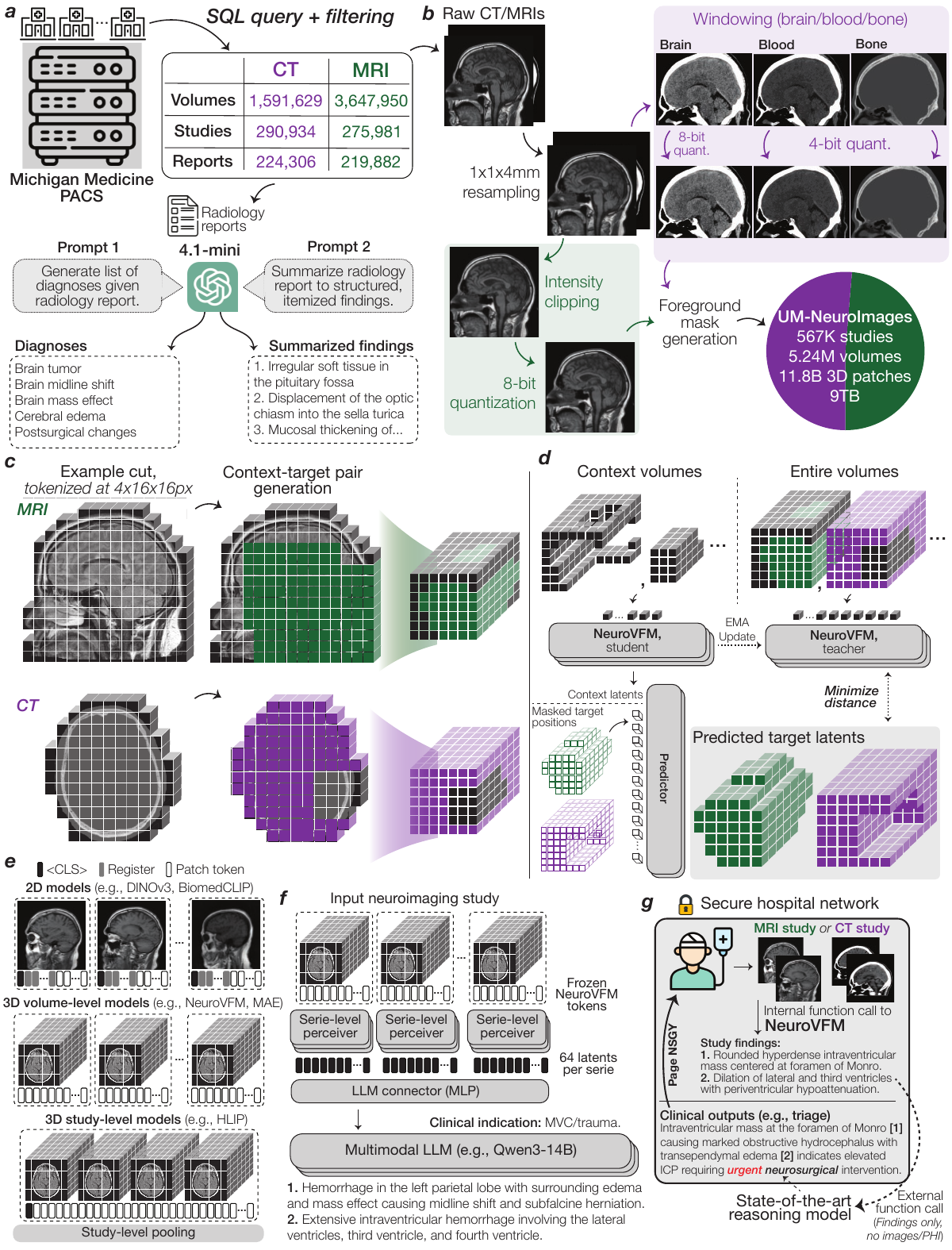}
    \caption{\textbf{Extended NeuroVFM workflow}. Caption on next page.}
    \label{exfig:ex_data1}
\end{figure*}

\begin{figure*}[p!]\ContinuedFloat
  \caption{\textbf{Extended NeuroVFM workflow}.
    \textbf{a}, We queried the Michigan Medicine PACS for neuroimaging studies of the head, brain, face, orbits, and neck. After removing non-image and corrupted entries, the UM-NeuroImages cohort comprises 566,915 CT and MRI studies, of which 444,188 have paired radiology reports. Per-modality counts of volumes, studies, and reports are shown. To obtain training and evaluation labels at scale, we used an LLM report-parsing pipeline \cite{Lyu2025-bt} that (1) extracts a list of expert-defined diagnoses and (2) converts free-text impressions to structured, itemized findings. Examples of both outputs are shown.
    \textbf{b}, Raw CT/MRI volumes are first resampled to \(1\times1\times4\,\mathrm{mm}\) (\(4\,\mathrm{mm}\) along the native acquisition axis) to reflect conventional clinical slice spacing. MRI intensities are clipped to 0.5-99.5th percentile, and CT volumes are windowed into brain, blood, and bone using standard Hounsfield windows. All volumes are quantized to 8-bit (CT blood/bone further reduced to 4-bit) to reduce footprint. Modality-specific foreground masks remove air only (no skull-stripping). The resulting dataset contains 11.8 billion 3D 4x16x16 patches (approximately 9 TB).
    \textbf{c}, Example CT/MRI volumes after background removal and tokenization. For Vol-JEPA training, volumes are partitioned into context-target pairs. Mask placements and lengths are randomized to cover diverse spatial configurations.
    \textbf{d}, In Vol-JEPA, a student volumetric transformer first encodes the context, after which a predictor combines these latents with the masked target positions to predict target latents. A teacher encoder (an EMA of the student) processes the full volume to produce target latents for the masked region. The objective minimizes the smooth L1 distance between the predicted and teacher target latents. Only the student and predictor receive gradients, and ultimately the student is used for downstream inference.
    \textbf{e}, To compare fairly across architectures, we use a unified study-level attentive probing strategy across all available tokens. 2D models (e.g., DINOv3 and BiomedCLIP) encode each slice, with classification, register, and patch tokens from all slices fed to the study-level probe. 3D volume-level models (e.g., NeuroVFM and VideoMAE) encode each volume, with all tokens pooled jointly. 3D study-level models (e.g., HLIP) natively output study classification tokens, but for parity we use all available output tokens in the study-level probe.
    \textbf{f}, We used NeuroVFM to condition a multimodal LLM in a LLaVa-1.5 style to generate preliminary neuroimaging findings. First, each volume in a study is encoded with frozen NeuroVFM. The resulting tokens are condensed by a per-volume Perceiver resampler to 64 latents. These then pass through the LLM connector (two-layer MLP) and are fed as vision tokens to a decoder-only LLM (e.g., Qwen3-14B). During finetuning, the clinical indication is given as text input and the model is trained to generate the itemized findings from \textbf{a}. An example indication and generated findings are shown.
    \textbf{g}, Illustrative deployment of NeuroVFM within a secure hospital environment. Each time a patient undergoes neuroimaging, an internal function call applies NeuroVFM to generate preliminary findings. An external state-of-the-art reasoning model receives only these findings (no images or PHI) to perform tasks such as triage and referral. Its output can then trigger service-line alerts (e.g., "page neurosurgery").}
\end{figure*}

\clearpage
\begin{figure*}[p!]
    \centering\includegraphics[scale=0.78]{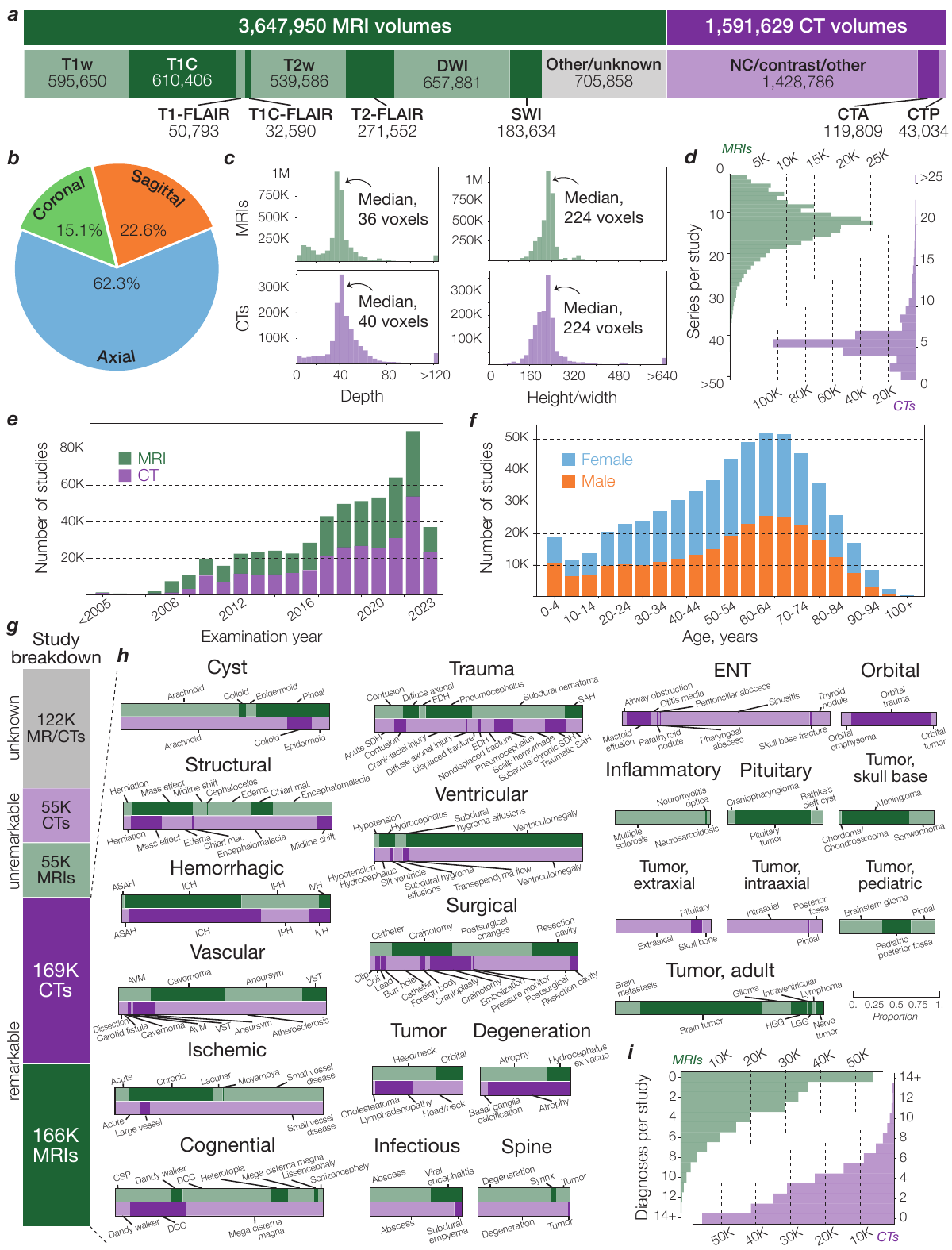}
    \caption{\textbf{Overview of UM-NeuroImages}. Caption on next page.}
    \label{exfig:ex_data2}
\end{figure*}

\begin{figure*}[p!]\ContinuedFloat
\caption{\textbf{Overview of UM-NeuroImages}. Descriptive statistics for the retrospective training corpus used for NeuroVFM.
\textbf{a}, Volume counts by modality and sequence family. Sequence families were inferred from the DICOM "Series Description" via keyword matching. Counts are therefore approximate and may not perfectly reflect acquisition parameters.
\textbf{b}, Distribution of volume orientation (axial, sagittal, coronal). 
\textbf{c}, Voxel-space distributions by modality: depth (slices) and in-plane dimension (height/width). Medians are indicated (MRI depth, 36 slices; CT depth, 40 slices; in-plane size for both, 224 voxels). 
\textbf{d}, Distribution of series per study by modality, showing a higher number of series in MRI (mode, 12) relative to CT (mode, 5).
\textbf{e}, Study counts by calendar year (2005-2023) for MRI and CT. June 1, 2023 marks the cut-off for inclusion in the training set.
\textbf{f}, Study counts by patient age (5-year bins) and sex.
\textbf{g}, Breakdown of studies by label status: remarkable (at least 1 positive diagnosis), unremarkable (no reported abnormality), and unknown (no paired report). Totals shown correspond to 168,579 CT and 165,780 MRI remarkable studies; 55,444 CT and 55,051 MRI unremarkable studies; and 122,061 studies with no retrieved report used for Vol-JEPA self-supervised pretraining.
\textbf{h}, Per-category prevalence of diagnoses (82 CT labels, 74 MRI labels). The aim was to design a clinically relevant ontology of diagnostic classes to evaluate our representation models. We refer to this benchmark as \emph{UM-NeuroBench}. Abbreviations: Hemorrhagic -- ASAH, aneurysmal subarachnoid hemorrhage; ICH, intracranial hemorrhage; IPH, intraparenchymal hemorrhage; IVH, intraventricular hemorrhage. Vascular -- AVM, arteriovenous malformation; VST, venous sinus thrombosis. Congenital -- CSP, cavum septum pellucidum; DCC, dysgenesis of the corpus callosum. Trauma -- EDH, epidural hematoma; SAH, subarachnoid hemorrhage; SDH, subdural hematoma. Tumor, adult -- HGG, high-grade glioma; LGG, low-grade glioma.
\textbf{i}, Number of positive diagnoses per study by modality. The long-tailed, multi-label distribution underscores our label breadth and enables training strong multi-diagnostic classifiers.}
\end{figure*}

\clearpage
\begin{figure*}[p!]
    \centering\includegraphics[scale=0.77]{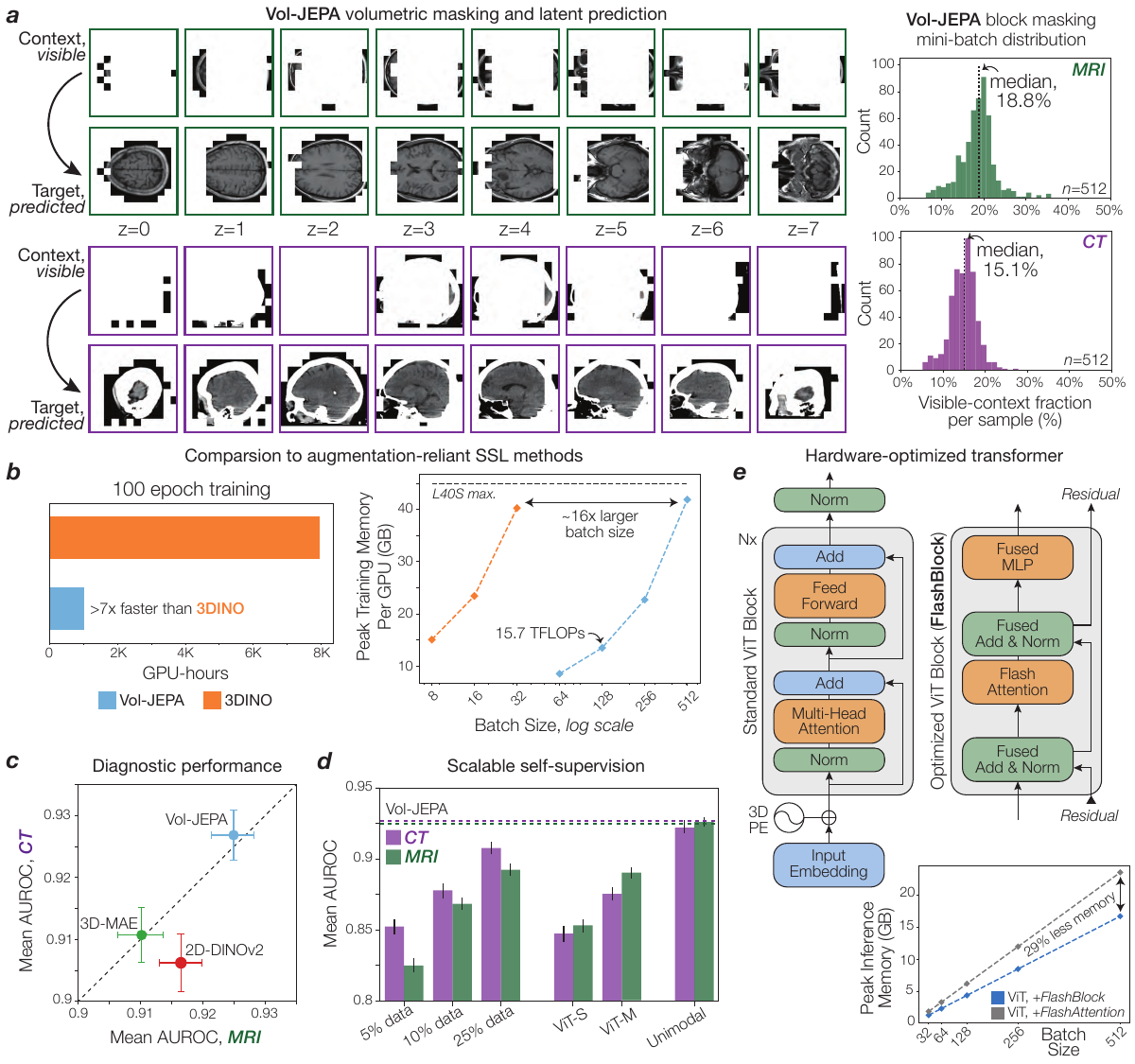}
    \caption{\textbf{Learning with Vol-JEPA}. Caption on next page.}
    \label{exfig:ex_data3}
\end{figure*}

\begin{figure*}[p!]\ContinuedFloat
    \caption{\textbf{Learning with Vol-JEPA}.
    \textbf{a}, For each Vol-JEPA training instance, we generate context-target pairs by sampling (1) one or more large blocks whose union forms the masked target, or (2) a single small block that serves as the visible context. These blocks span all three axes, ensuring the encoder sees a diverse set of spatial volumes. Examples are shown for MRI and CT volumes across eight axial slices (z=0-7). 
    \textbf{b}, Mini-batch  histograms (n=512) show the per-instance visible-context fraction (MRI median, 18.8\%; CT median, 15.1\%). CT uses slightly smaller contexts because out-of-body voxels are reliably background. Since MRI foregrounds often include air due to lack of explicit skull stripping, we keep a slightly larger visible fraction to preserve anatomical signal. The range of fractions encourages the encoder to model shared neuroanatomy rather than background shortcuts.
    \textbf{c}, Robust volumetric self-supervised learning is typically expensive due to heavy augmentations and large forward passes. Left, a 3D-DINO-style baseline using MONAI 3D augmentations requires >7x GPU-hours than Vol-JEPA to train for 100 epochs. Right, peak training memory per GPU vs. batch size (log scale) shows Vol-JEPA encoding 16x larger batches for any given memory below the L40S limit. At batch size 128, throughput is approximately 15.7 TFLOPs/GPU.
    \textbf{d}, We trained a compute-matched VideoMAE (85\% random masking) and DINOv2 on 2D images (UM-NeuroImages training corpus has 209M unique slices). Using the unified study-level attentive probing protocol (Extended Data Figure 1e), Vol-JEPA attains the highest mean AUROC on UM-NeuroBench for MRI and CT, indicating advantages of masked volumetric latent prediction over decoding-centric (VideoMAE) and 2D slice-only (DINOv2) approaches.
    \textbf{e}, We replaced the standard pre-norm ViT block (with FlashAttention-2) with a FlashAttention-2 block with operator fusion (Tri Dao, 2023). The schematic provided contrasts both implementations. Benchmarking demonstrated 29\% lower peak inference memory used across batch sizes for ViT+FlashBlock vs. a standard ViT+FlashAttention stack, facilitating on-premises hospital deployment.}
\end{figure*}

\clearpage
\begin{figure*}[p!]
    \centering\includegraphics[scale=0.8]{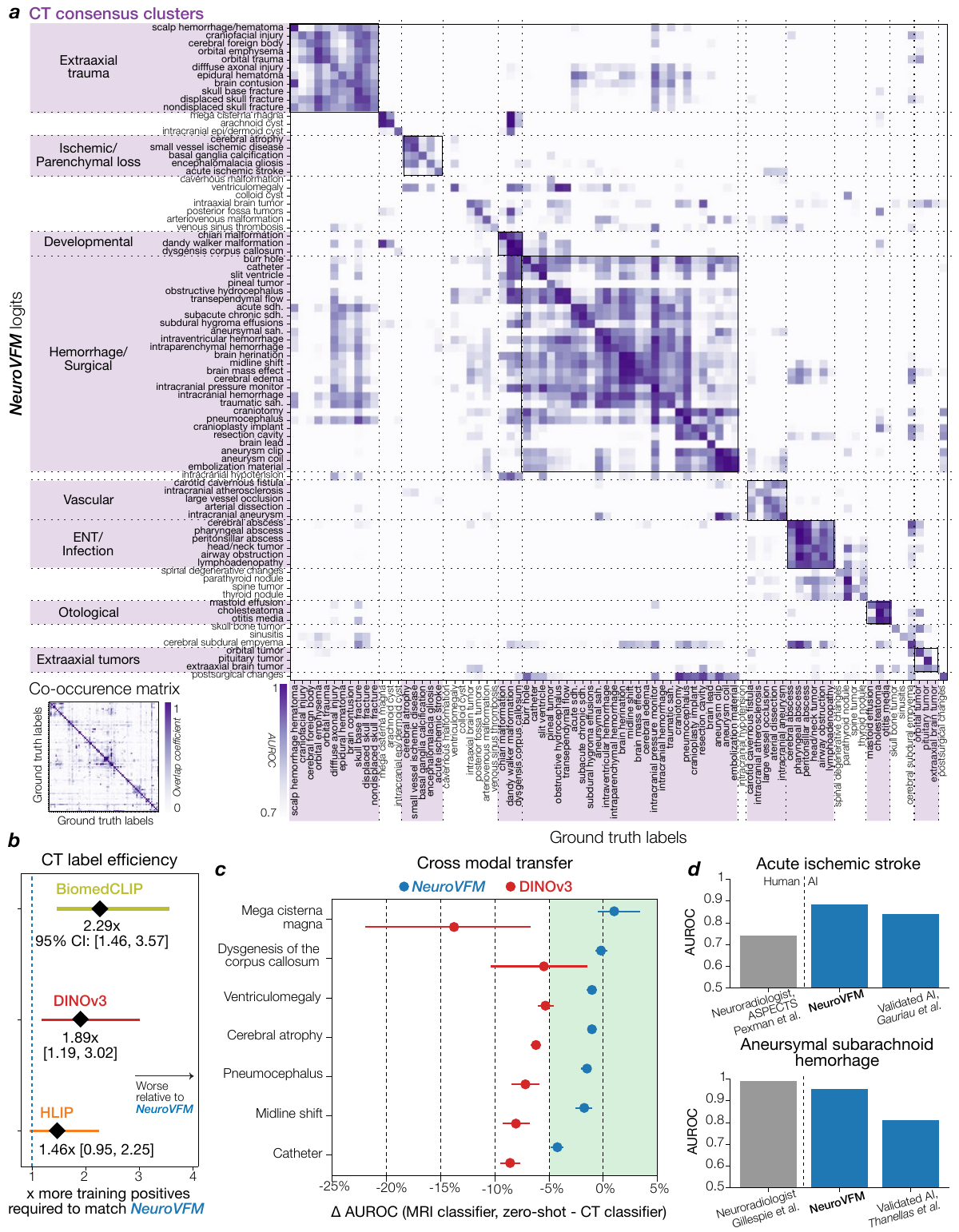}
    \caption{\textbf{Extended CT diagnostic results}. Caption on next page.}
    \label{exfig:ex_data4}
\end{figure*}

\begin{figure*}[p!]\ContinuedFloat
    \caption{\textbf{Extended CT diagnostic results}. CT diagnostic evaluation comprises 82 diagnoses defined on UM-NeuroImages. All encoders were frozen and evaluated using the same study-level attentive probe (Extended Data Fig. \ref{exfig:ex_data1}e).
    \textbf{a}, Logit-label performance matrix for the NeuroVFM CT classifier, where each cell is AUROC of $\text{logit}_{i}$ against ground-truth $\text{label}_j$ (main diagonal shows per-diagnosis AUROC). Rows and columns are reorganized by hierarchical consensus clustering (bootstrapped similarity, average linkage). Lower left, reference co-occurrence matrix computed from ground truth using min-normalization, $M_{ij} = \dfrac{|i \cap j|}{\min\!\left(|i|,\,|j|\right)}$.
    \textbf{b}, Following the empirical scaling observed in Fig. \ref{fig:fig_2_results}e, we fit a shared slope across NeuroVFM and three baselines (BiomedCLIP, DINOv3, and HLIP) via ANCOVA (\(p\ll 0.05\)), relating calibrated F1 score to \(\log_{10}\) of the number of positive training instances. Intercept shifts yield data-equivalence factors (how many more positives a baseline requires to match NeuroVFM at a fixed F1), shown with 95\% CIs.
    \textbf{c}, Cross-modal transfer results for seven diagnoses with shared semantics across MRI/CT. An MRI-trained classifier is evaluated \emph{zero-shot} on CT, and we plot \(\Delta\)AUROC = (MRI-on-CT) $-$ (CT-on-CT) with 95\% bootstrap CIs. For NeuroVFM, all differentials lie within the a priori equivalence band of \(\pm 0.05\) (green), whereas DINOv3 exhibits consistent modality shift.
    \textbf{d}, We compared NeuroVFM results on two representative CT detection tasks (top, acute ischemic stroke; bottom, aneurysmal subarachnoid hemorrhage) to published studies. NeuroVFM achieves AUROC comparable to expert neuroradiologists \cite{Pexman2001-zt, Gillespie2025-kq} and to validated AI systems \cite{Gauriau2023-lq, Thanellas2023-fq}. For prior studies, AUROC was approximated as the mean of the reported sensitivity and specificity.}
\end{figure*}

\clearpage
\begin{figure*}[p!]
    \centering\includegraphics[width=\textwidth]{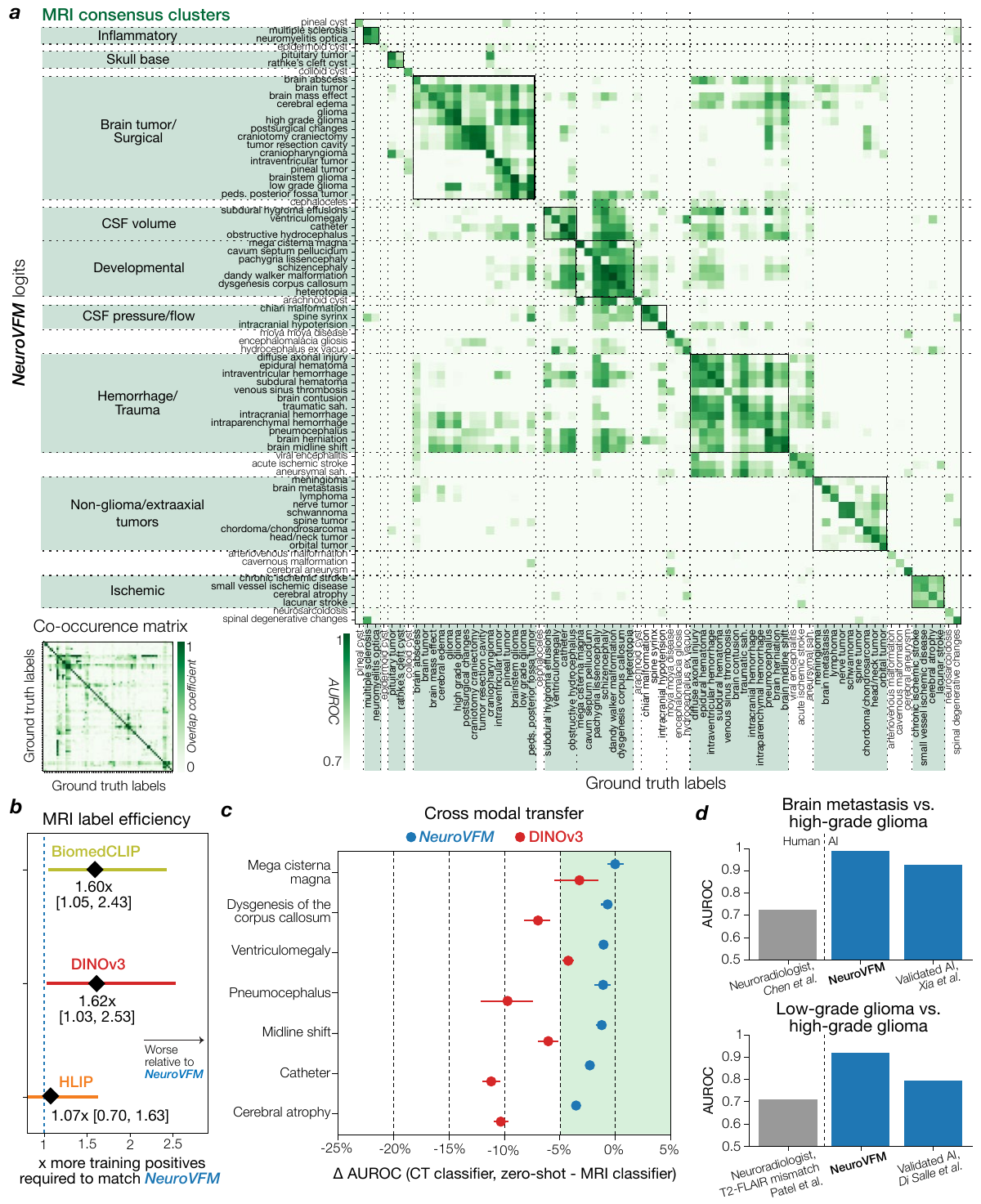}
    \caption{\textbf{Extended MRI diagnostic results. Caption on next page.}}
    \label{exfig:ex_data5}
\end{figure*}

\begin{figure*}[p!]\ContinuedFloat
    \caption{\textbf{Extended MRI diagnostic results}. MRI diagnostic evaluation comprises 74 diagnoses defined on UM-NeuroImages. All encoders were frozen and evaluated using the same study-level attentive probe (Extended Data Fig. \ref{exfig:ex_data1}e).
    \textbf{a}, Logit-label performance matrix for the NeuroVFM MRI classifier, where each cell is AUROC of $\text{logit}_{i}$ against ground-truth $\text{label}_j$ (main diagonal shows per-diagnosis AUROC). Rows and columns are reorganized by hierarchical consensus clustering (bootstrapped similarity, average linkage). Lower left, reference co-occurrence matrix computed from ground truth using min-normalization, $M_{ij} = \dfrac{|i \cap j|}{\min\!\left(|i|,\,|j|\right)}$.
    \textbf{b}, Following the empirical scaling observed in Fig. \ref{fig:fig_2_results}e, we fit a shared slope across NeuroVFM and three baselines (BiomedCLIP, DINOv3, and HLIP) via ANCOVA (\(p\ll 0.05\)), relating calibrated F1 score to \(\log_{10}\) of the number of positive training instances. Intercept shifts yield data-equivalence factors (how many more positives a baseline requires to match NeuroVFM at a fixed F1), shown with 95\% CIs.
    \textbf{c}, Cross-modal transfer results for seven diagnoses with shared semantics across MRI/CT. A CT-trained classifier is evaluated \emph{zero-shot} on MRI, and we plot \(\Delta\)AUROC = (CT-on-MRI) $-$ (MRI-on-MRI) with 95\% bootstrap CIs. For NeuroVFM, all differentials lie within the a priori equivalence band of \(\pm 0.05\) (green), whereas DINOv3 exhibits consistent modality shift.
    \textbf{d}, We compared NeuroVFM results on two representative MRI differentiation tasks (top, brain metastasis vs. high-grade glioma; bottom, low- vs. high-grade glioma) to published studies. For NeuroVFM, the score was the pairwise logit margin ($\text{logit}_1$-$\text{logit}_2$), and AUROC was computed on this margin. NeuroVFM achieves AUROC comparable to expert neuroradiologists \cite{Patel2017-pm, Chen2012-nf} and to validated AI systems \cite{Di-Salle2024-cp, Xia2025-te}. For prior studies that only report sensitivity and specificity, we approximate the AUROC as the average.}
\end{figure*}
\clearpage

\begin{figure*}[p!]
    \centering\includegraphics[width=\textwidth]{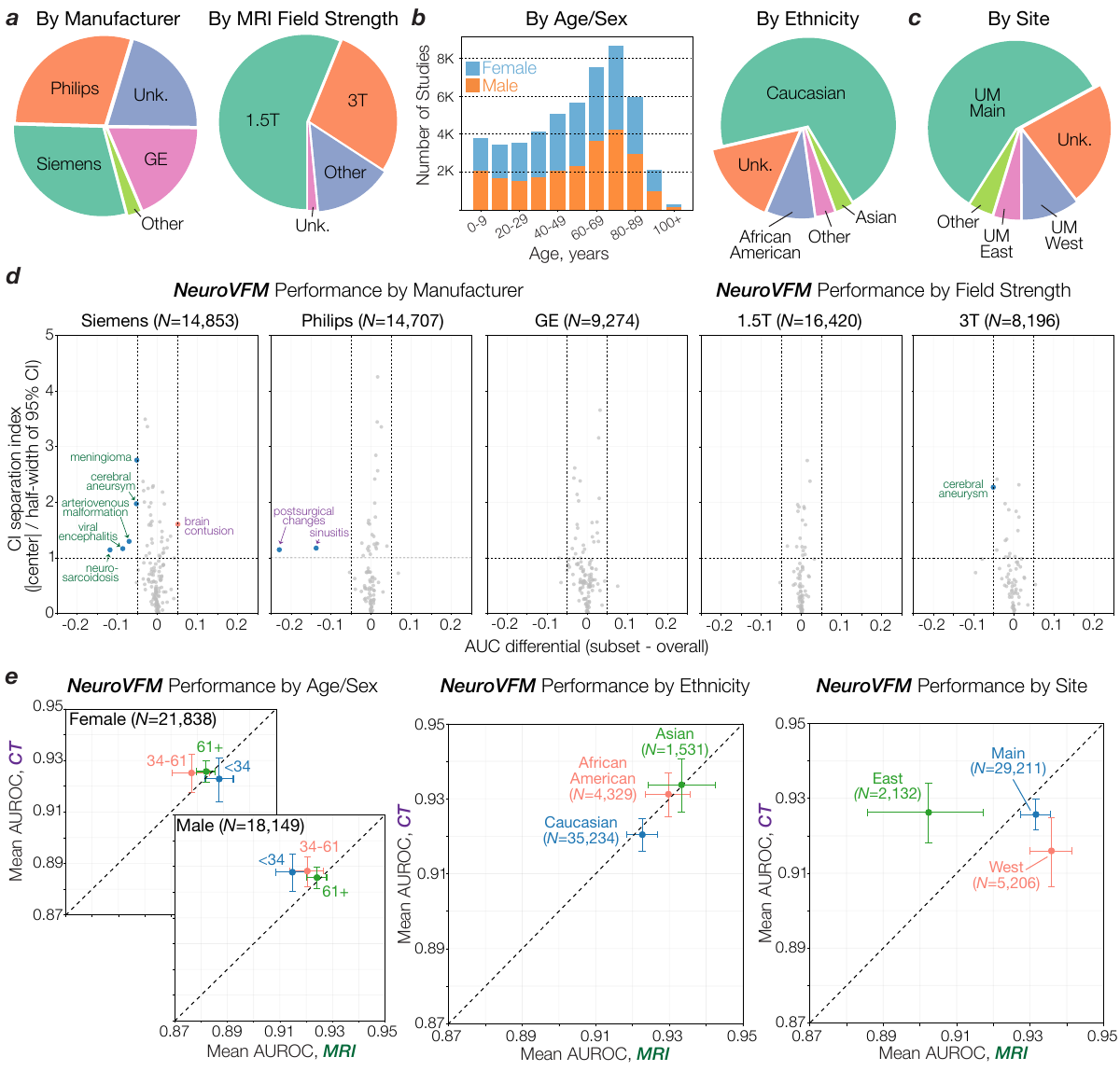}
    \caption{\textbf{Subgroup and health system analysis.} Descriptive statistics for the prospective testing corpus and robustness of NeuroVFM across common sources of health system variation.
    \textbf{a}, Composition by scanner manufacturer and MRI field strength.
    \textbf{b}, Composition by age/sex and ethnicity.
    \textbf{c}, Composition by acquisition site within the health system. We limited the analysis to the three most common sites: University of Michigan (UM) Main, West, and East.
    \textbf{d}, Per-diagnosis robustness checks by (left) manufacturer and (right) MRI field strength. For each eligible cell (subgroup $\times$ diagnosis; requires at least 10 positives and negatives present), we compute the AUC differential relative to the full test set ($\Delta_g=\mathrm{AUC}_g-\mathrm{AUC}_{\text{all}}$) using 10,000 paired, study-level bootstrap resamples to form percentile 95\% CIs. Points show $\Delta_g$, and the vertical dashed lines mark an a priori equivalence band of $\pm$ 0.05 AUROC. The y-axis summarizes CI separation index, $S = \frac{\Delta_g}{\text{half-width}_{\text{95\%}}}$, where $S>1$ indicates the 95\% CI excludes 0 (i.e., a significant deviation). Cells with 95\% CIs entirely within $\pm$ 0.05 AUROC were interpreted as showing no material deviation. Most diagnoses remain within the equivalence band, with outliers annotated (CT, purple; MRI, green).
    \textbf{e}, Mean AUROC ($\pm$ 95\% CIs) for MRI vs. CT stratified by age/sex, ethnicity, and site. No material differences present for age/sex and ethnicity. At the site level, MRI performance is lower at the East site, while CT is comparable across all.}
    \label{exfig:ex_data6}
\end{figure*}

\begin{figure*}[p!]
    \centering\includegraphics[width=\textwidth]{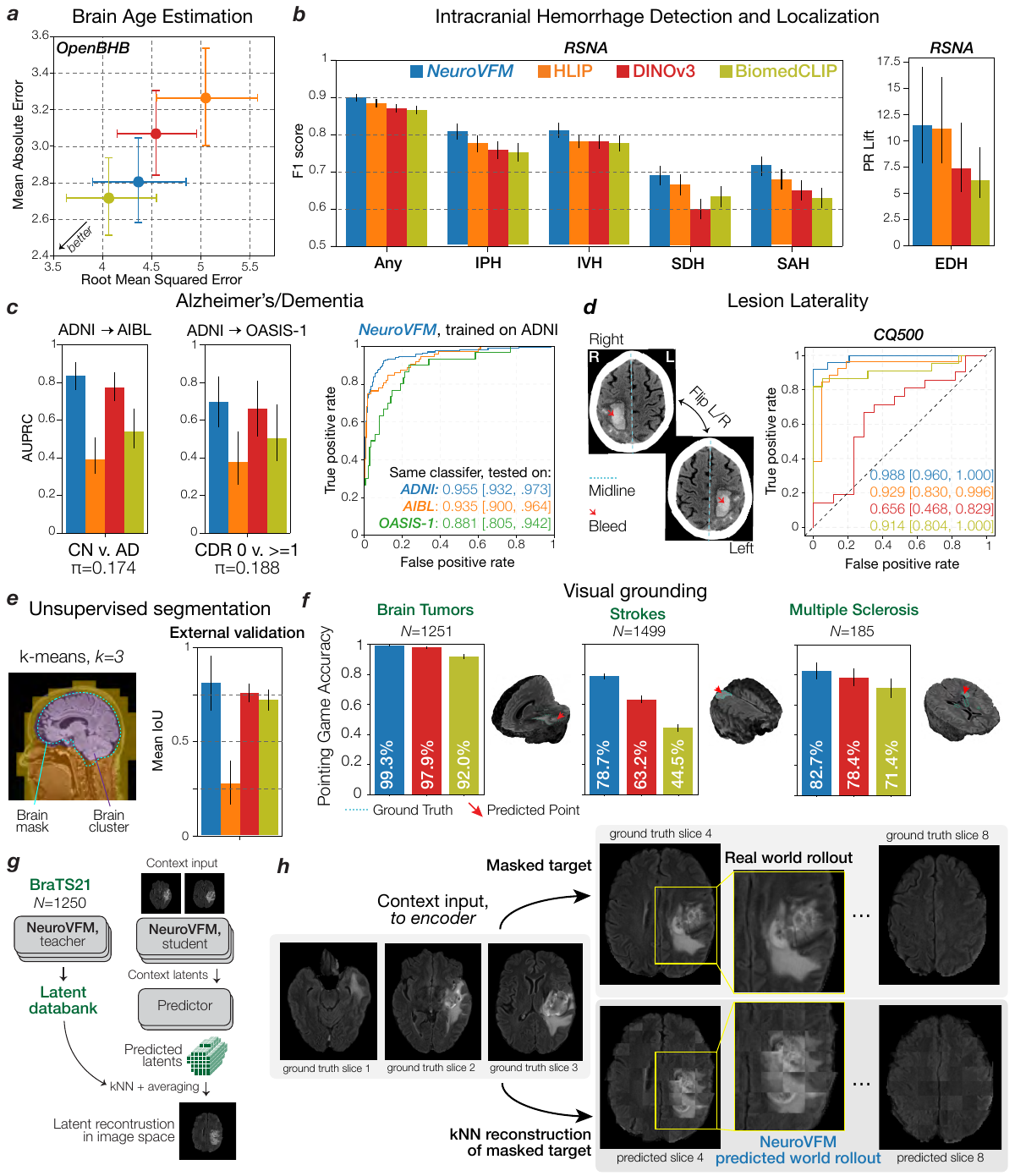}
    \caption{\textbf{Open benchmarks performance}. Caption on next page.}
    \label{exfig:ex_data7}
\end{figure*}

\begin{figure*}[p!]\ContinuedFloat
    \caption{\textbf{Open benchmarks performance.} All encoders (NeuroVFM, blue; HLIP, orange; DINOv3, red; BiomedCLIP, olive) were frozen and evaluated using the same study-level attentive probe (Extended Data Fig. \ref{exfig:ex_data1}e). Error bars show 95\% CIs.
    \textbf{a}, Brain age estimation results on the OpenBHB dataset (T1 MRIs). Scatter of RMSE (x-axis) vs. MAE (y-axis) per model shown. NeuroVFM is competitive (MAE, 2.805 years), second to BiomedCLIP. Strong 2D baselines likely benefit from dense slice-level features available to the probe.
    \textbf{b}, Intracranial hemorrhage detection and localization results on the 2019 RSNA-ICH dataset (non-contrast CT). Left, F1 for ICH detection and each subtype (IPH, IVH, SDH, SAH) shown. Right, PR lift for EDH (AUPRC/prevalence; EDH, 1.88\%). NeuroVFM matches or exceeds other baselines across all tasks.
    \textbf{c}, We used three Alzheimer's/dementia cohorts to test whether features generalize beyond a single study: ADNI (training, with a held-out test set) and two independent test sets (AIBL and OASIS-1). ADNI to AIBL reflects classic domain shift, whereas ADNI to OASIS-1 probes robustness as OASIS-1 provides a dementia severity score (CDR 0 vs. $\geq$1). Left, training on ADNI and testing on AIBL/OASIS-1 yields strong AUPRC for NeuroVFM (DINOv3 comparable). Right, the same ADNI-trained NeuroVFM classifier exhibits high AUROC on the held-out ADNI test set and transfers well to AIBL and OASIS-1. These results indicate the learned NeuroVFM features capture disease-relevant information rather than site-specific cues, consistent with Fig. \ref{fig:fig_3_scene}e where attention concentrates on structures known to be discriminative for Alzheimer's disease \cite{Planche2022-ps}.
    \textbf{d}, Correct laterality assignment is clinically important yet rarely tested. The CQ500 dataset provides expert "Bleed-Left/Bleed-Right" labels, allowing a targeted laterality check given correct bleed detection. For each model, among studies it correctly flags as bleed, we horizontally flip the volume and measure changes in both the Bleed-Left and Bleed-Right logits (i.e., $\Delta_L = \text{logit}_L^{\text{flipped}} - \text{logit}_L, \Delta_R = \text{logit}_R^{\text{flipped}} - \text{logit}_R$). We summarize performance as AUROC over $\Delta_L$ and $\Delta_R$ against the ground-truth side. NeuroVFM is near perfect, representative of strong anatomical encoding. DINOv3, on the other hand, underperforms, plausibly due to strong flip-invariance from augmentations.
    \textbf{e}, K-means clustering (k=3) of dense patch embeddings from NeuroVFM yields coarse tissue-wise clusters that recover brain parenchyma without supervision. See Methods for full description.
    \textbf{f}, We quantitatively validated attention maps of models' UM-NeuroImages classifier against ground-truth segmentations on the BraTS21 ('brain tumor'), ATLAS v2.0 ('acute ischemic stroke'), and OpenMS ('multiple sclerosis') datasets. Specifically, we computed whether the voxel with the highest attention lies in the ground-truth segmentation, known as \textit{pointing game accuracy}. Result and a representative example for each dataset are shown, with NeuroVFM demonstrating the highest degree of visual grounding.
    \textbf{g}, Schematic outlining how the predictor of NeuroVFM can be visualized via kNN reconstruction. To probe what the predictor infers for a masked target region, we project predicted latents back to image space via $k$-nearest-neighbor (kNN) retrieval. First, we build a patch latent databank by encoding patches from a reference set with the EMA teacher, yielding latent keys $z_i$ and storing their corresponding image patches $x_i$. For a held-out volume, we encode a context region with the student encoder and use the predictor to produce latent predictions $\hat{z}_j$ for each masked target patch. For each $\hat{z}_j$, we perform kNN search (cosine distance) over $\{z_i\}$, retrieve the top-$k$ patches, and form a pseudo-reconstruction $\tilde{x}_j = \frac{1}{k}\sum_{i\in N_k}x_i$. 
    \textbf{h}, Example rollout of a FLAIR volume from BraTS21. Here, three axial slices were passed to the student as context and the rest of the sequence was predicted and rolled out. Qualitatively, the reconstructions recover tumor extent and surrounding anatomy from the correct neuroimaging sequence and rollouts across adjacent slices maintain structural consistency, indicating that the predicted latents are accurate.
    }
\end{figure*}

\clearpage
\begin{figure*}[p!]
    \centering\includegraphics[scale=0.80]{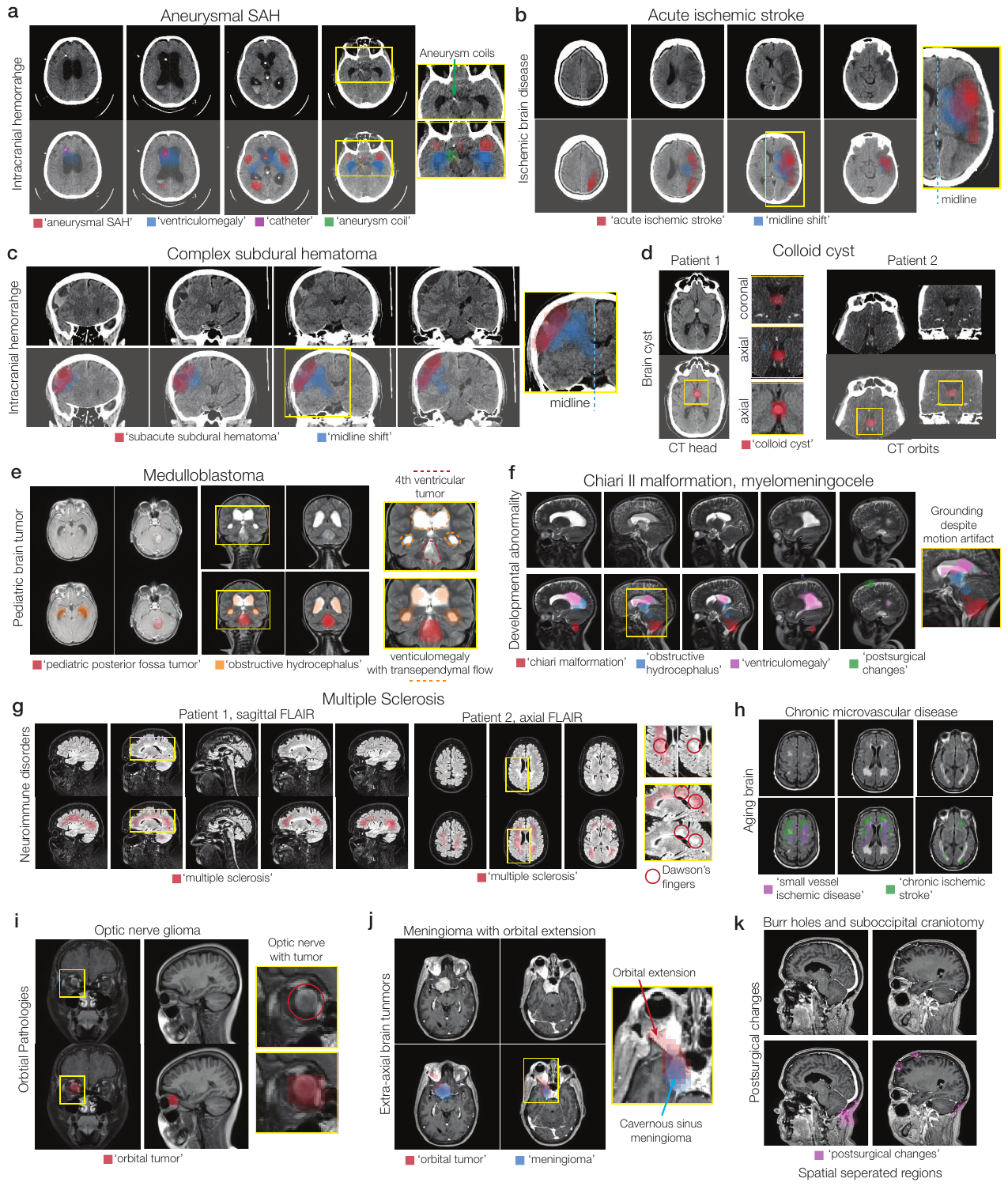}
    \caption{\textbf{Neuroimage scene understanding and grounded diagnoses.} Caption on next page.}
    \label{exfig:ex_data8}
\end{figure*}

\begin{figure*}[p!]\ContinuedFloat
    \caption{\textbf{Neuroimage scene understanding and grounded diagnoses.} A diverse set of illustrative examples that demonstrate NeuroVFM is a grounded visual foundation model. 
    \textbf{a}, An example of aneurysmal subarachnoid hemorrhage, treated via endovascular coiling, with associated hydrocephalus and ventricular catheter. NeuroVFM can parse the head CT into diagnostic regions and associate correct diagnoses. \textbf{b}, A patient with an acute middle cerebral artery ischemic stroke and associated cerebral edema and midline shift. \textbf{c}, A patient with a complex, septated, mixed-density subdural hematoma also causing midline shift. Note that NeuroVFM identifies midline shift regardless of the underlying etiology. \textbf{d}, Two patients with classic third ventricle colloid cysts. NeuroVFM attends to the colloid cyst regardless of patient orientation or imaging protocol. \textbf{e}, An example of a pediatric posterior fossa tumor causing obstructive hydrocephalus. NeuroVFM performs well across age groups. \textbf{f}, Example of a Chiari II malformation with associated obstructive hydrocephalus, ventriculomegaly, and previous cerebrospinal shunt placement.  \textbf{g}, Two patients with newly diagnosed multiple sclerosis. Both patients have classic periventricular demyelination, known as Dawson's fingers, which NeuroVFM identifies on both sagittal and axial orientations. \textbf{h} NeuroVFM can differentiate between multiple sclerosis and periventricular white matter changes due to chronic microvascular disease. An example of an (\textbf{i}) optic nerve glioma and a (\textbf{j}) cavernous sinus meningioma with intraorbital extension. NeuroVFM performance generalizes well to ophthalmologic conditions. \textbf{k}, NeuroVFM can associate spatially separated regions, for example, surgeries completed in different locations, with the same underlying diagnosis. NeuroVFM can identify postsurgical changes throughout the neuroimaging study.}
\end{figure*}
\clearpage

\begin{figure*}[p!]
    \centering\includegraphics[scale=0.77]{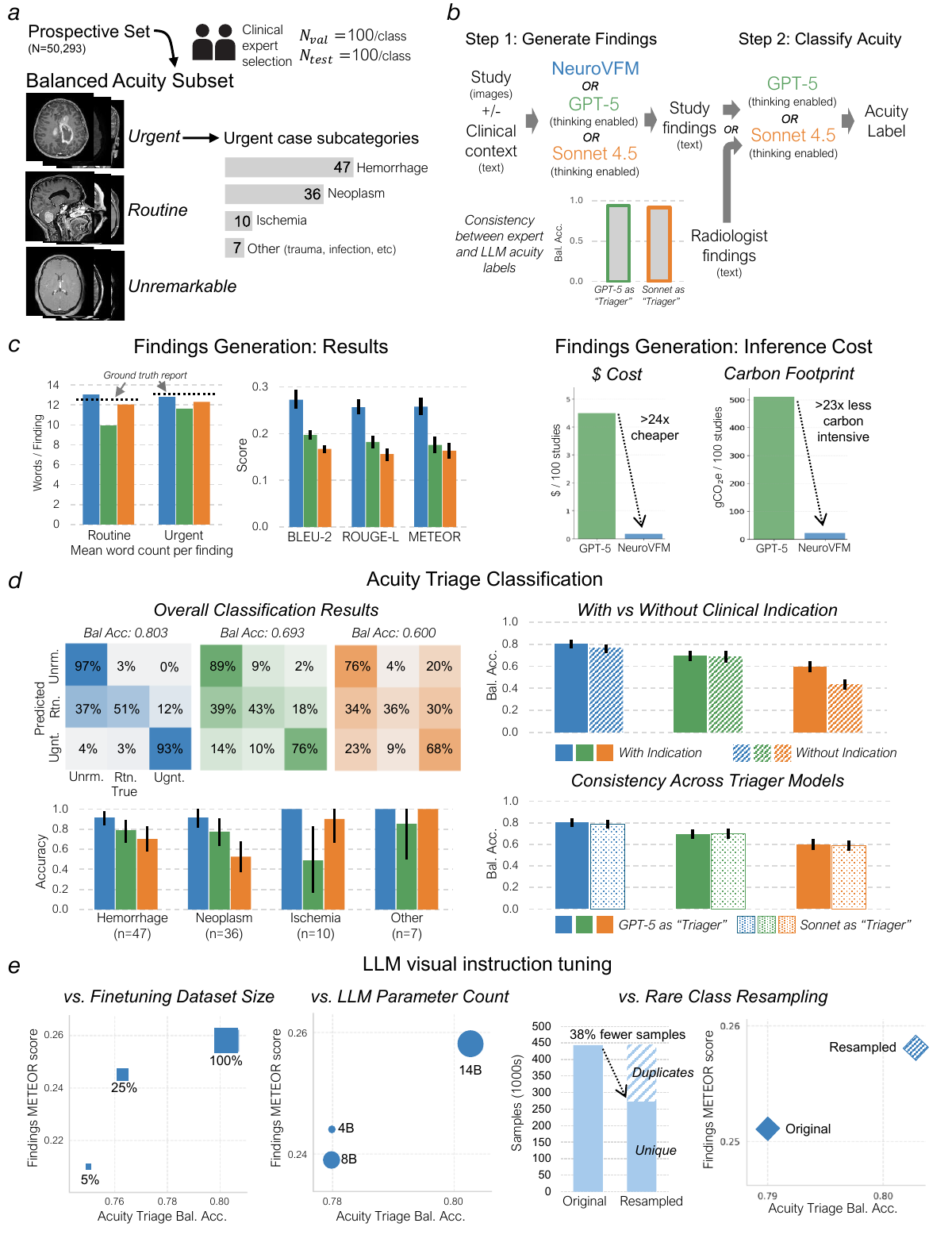}
    \caption{\textbf{Generative evaluation}. Caption on next page.}
    \label{exfig:ex_data9}
\end{figure*}

\begin{figure*}[p!]\ContinuedFloat
    \caption{\textbf{Generative evaluation.}
    \textbf{a}, Composition of the 'UM-NeuroImages-Triage' evaluation dataset. The set consists of 600 studies (300 validation, 300 holdout) derived from our prospective test set. This set was hand-selected by neuroimaging experts to be balanced across three acuity classes ("Unremarkable", "Routine", "Urgent") and two modalities (MRI, CT). \textbf{b}, Schematic of the two-stage evaluation pipeline for neuroimaging report generation and clinical triage. Step 1 (Generate Findings): 3D studies (for NeuroVFM) or converted 2D slices (for GPT-5, Claude Sonnet 4.5) are provided to each model with a one-sentence clinical indication to generate key findings. Step 2 (Classify Acuity): The generated text is passed to an "LLM-as-a-judge" (Triager LLM: GPT-5 or Claude Sonnet 4.5) to classify the study into an acuity class. The pipeline was validated by passing ground-truth radiologist findings to the Triager LLMs, which achieved >94\% accuracy in replicating the expert-designated acuity labels. \textbf{c}, Generated report quality and inference cost analysis. Left: Bar plots comparing NeuroVFM, GPT-5, and Claude Sonnet 4.5 on traditional NLP metrics (BLEU-2, ROUGE-L, METEOR) against ground-truth summarized reports. Right: Post-hoc analysis of inference efficiency, comparing dollar cost (USD) and carbon footprint (gCO2e) per 100 studies. NeuroVFM estimates assume inference on a single NVIDIA L40S GPU (AWS G6e instance) with a 3s/study inference time and southeast Michigan power grid intensity (0.5869 kgCO2e/kWh). GPT-5 estimates use October 2025 batch API pricing and carbon estimates from Jegham et al. (medium reasoning, >10k input/>1k output tokens) \cite{Jegham2025-fp}. \textbf{d}, Clinical triage accuracy and ablation studies. Top Left: Triage accuracy (3-class) of findings generated by NeuroVFM, GPT-5, and Claude Sonnet 4.5, evaluated by the "LLM-as-a-judge" pipeline. Bottom Left: Triage accuracy stratified by key acute pathologies (e.g., hemorrhage, aggressive mass, ischemia). Top Right: Ablation study on the impact of withholding the 1-sentence clinical indication from the input prompt. Bottom Right: Ablation study assessing the robustness of triage accuracy when using different "LLM-as-a-judge" models (GPT-5 vs. Claude Sonnet 4.5). \textbf{e}, Scaling properties of NeuroVFM visual instruction tuning. Left: Model performance (METEOR score and Triage Accuracy) as a function of finetuning training dataset size. Center: Model performance as a function of LLM parameter count (Qwen3-4B, 8B, 14B). Right: Comparison of model performance between the final model (trained on a dataset resampled based on diagnostic class rarity) and a baseline model trained on the original, imbalanced dataset, demonstrating the performance gains from data curation.
    }
\end{figure*}
\clearpage

\begin{figure*}[p!]
    \centering\includegraphics[scale=0.75]{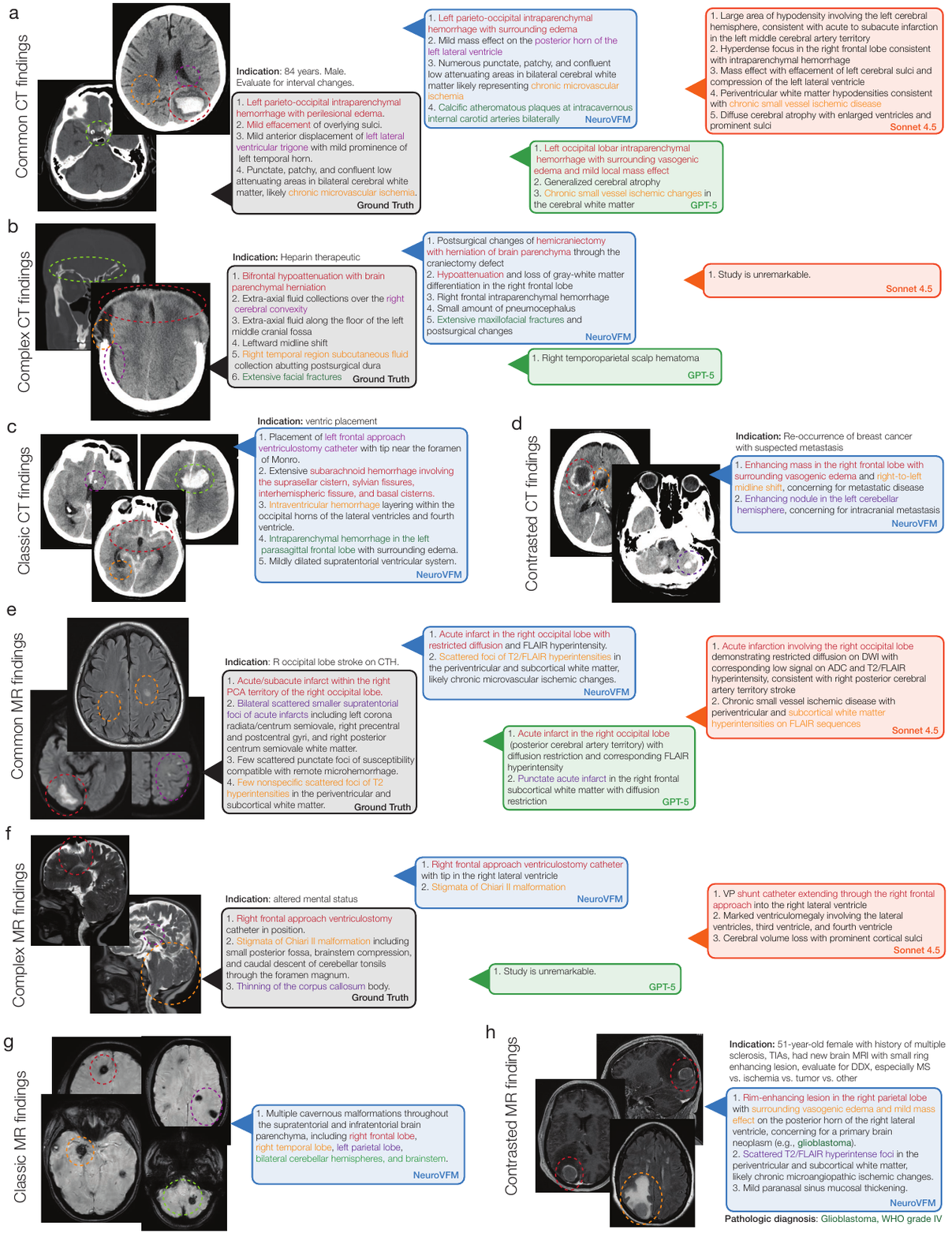}
    \caption{\textbf{Generated triage reports}. Caption on next page.}
    \label{exfig:ex_data10}
\end{figure*}

\begin{figure*}[p!]\ContinuedFloat
    \caption{\textbf{Generated triage reports.} A diverse set of illustrative examples of generated triage reports. 
    \textbf{a}, Example of a patient with a common lobar hemorrhage within the left parietal and occipital lobes. NeuroVFM and GPT-5 identified the hemorrhage and correctly localized it. \textbf{b}, A complex head CT of a patient who underwent a craniectomy after a traumatic brain injury. This example illustrates the challenge of interpreting \emph{real-world clinical} neuroimaging studies that do not fit neatly into singular diagnoses, which is often observed in standardized research datasets. NeuroVFM was able to recognize the postsurgical and post-traumatic changes, while both GPT-5 and Claude Sonnet 4.5 were not able to interpret this image correctly. \textbf{c}, Classic subarachnoid hemorrhage pattern due to a ruptured anterior communicating artery aneurysm. \textbf{d}, Contrasted CTs are much less common. NeuroVFM was able to both recognize that this was a contrasted CT and interpret that the findings were most consistent with metastatic brain disease. \textbf{e}, A common example of restricted diffusion due to an acute infarct located in the right posterior cerebral artery (PCA) distribution. \textbf{f}, A complex patient with Chiari II malformation and myelomeningiocele. NeuroVFM was trained on pediatric MRIs with developmental abnormalities and is, therefore, able to correctly interpret this MRI. \textbf{g}, Classic appearance of cerebral cavernous malformations. \textbf{h}, A patient with a complex medical history and indication has a new brain lesion. NeuroVFM correctly identified this tumor as most likely a glioblastoma, confirmed on pathologic diagnosis.
    }
\end{figure*}
\clearpage

\renewcommand{\figurename}{Supplementary Data Figure}
\setcounter{figure}{0}

\begin{figure*}[p!]
    \centering\includegraphics[scale=0.75]{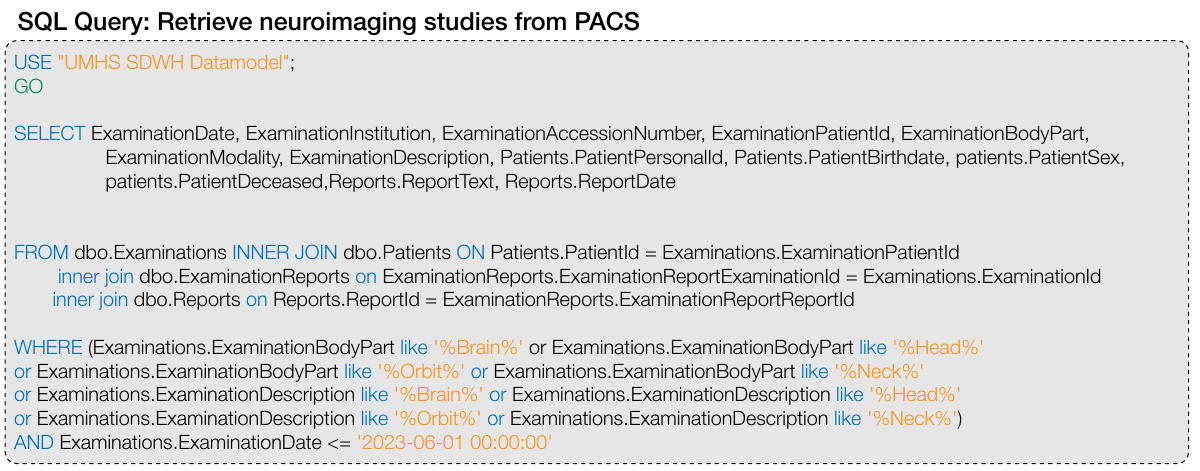}
    \caption{\textbf{SQL query to retrieve neuroimaging studies}. Using the SDW, we retrieved all CT and MRI studies whose body part or description contained “brain,” “head,” “orbit(s),” or “neck,” returning exam metadata and report text. The SQL query filters studies acquired before June 1 2023 for model development. The held-out prospective set applies the same query on and after that date.}
    \label{exfig:supp_data1}
\end{figure*}
\clearpage

\begin{figure*}[p!]
    \centering\includegraphics[scale=0.75]{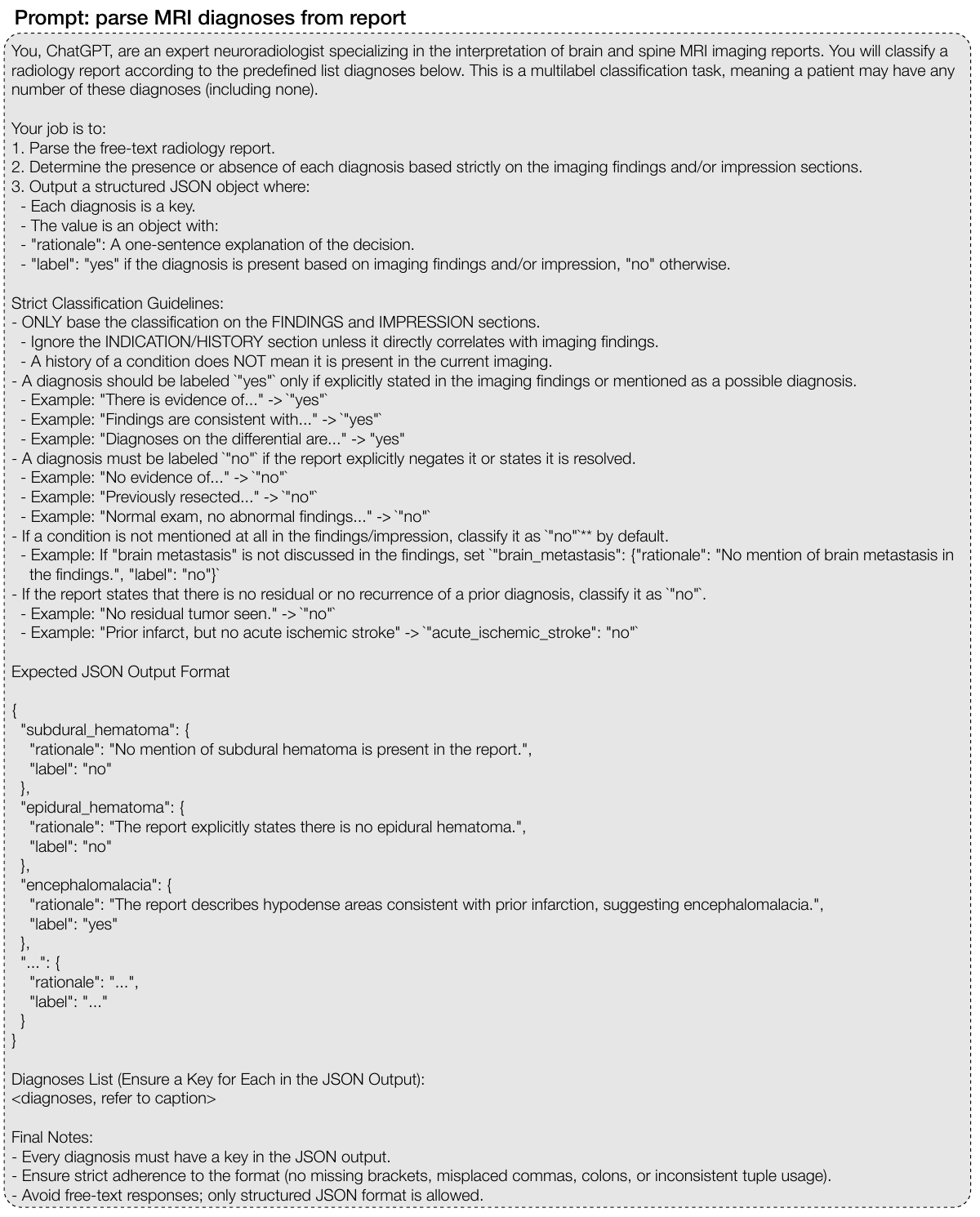}
    \caption{\textbf{MRI diagnosis extraction prompt.} Caption on next page.}
    \label{exfig:supp_data2}
\end{figure*}

\begin{figure*}[p!]\ContinuedFloat
    \caption{\textbf{MRI diagnosis extraction prompt.} We used an LLM-based annotation pipeline \cite{Lyu2025-bt} with GPT-4.1-mini to convert free-text radiology reports into structured labels for 74 expert-defined MRI diagnoses. The prompt requires, for each diagnosis, a present/absent label and a brief rationale supporting the decision. The full list of MRI diagnoses is: subdural\_hematoma; epidural\_hematoma; brain\_contusion; diffuse\_axonal\_injury; cerebral\_aneurysm; arteriovenous\_malformation; cavernous\_malformation\_cavernoma (``yes'' if radiologist confident for cavernoma; ``no'' if possible microbleed); venous\_sinus\_thrombosis; acute\_ischemic\_stroke (``yes'' for acute ischemia/infarct with diffusion restriction); chronic\_ischemic\_stroke (``yes'' for chronic or evolving ischemia/infarct without diffusion restriction); moya\_moya\_disease; small\_vessel\_ischemic\_disease (also known as microvascular ischemic disease); lacunar\_stroke; intracranial\_hemorrhage (``yes'' for any intracranial hemorrhage, excluding microhemorrhage); intraparenchymal\_hemorrhage (``yes'' for hemorrhages within the brain, usually spontaneous and due to hypertension); intraventricular\_hemorrhage; aneurysmal\_subarachnoid\_hemorrhage (``yes'' if due to ruptured aneurysm or other vascular cause); traumatic\_subarachnoid\_hemorrhage (``yes'' if due to head trauma); multiple\_sclerosis; neuromyelitis\_optica; neurosarcoidosis; brain\_abscess (``yes'' if radiologist is confident for brain abscesses or subdural empyemas); viral\_encephalitis (most commonly herpes encephalitis involving the temporal and frontal lobes); arachnoid\_cyst; pineal\_cyst; epidermoid\_cyst (``yes'' for intracranial epidermoid cyst only); colloid\_cyst; brain\_tumor (``yes'' for any intracranial mass, growth, or neoplasm); orbital\_tumor (``yes'' for any orbital mass, growth, or neoplasm); head\_neck\_tumor (``yes'' for tumor/cancer in the oral cavity, throat, larynx, nasal cavity, salivary, and thyroid gland); spine\_tumor (``yes'' for any spine mass, growth, or neoplasm); nerve\_tumor (``yes'' for nerve, nerve root, or nerve sheath tumors); spinal\_degenerative\_changes; glioma (``yes'' if mostly likely primary glial tumor); high\_grade\_glioma (``yes'' if MRI features for high-grade glioma); low\_grade\_glioma (``yes'' if MRI features for low-grade glioma); brain\_metastasis; meningioma; schwannoma; lymphoma; pineal\_tumor (``yes'' for any tumor in the pineal region); intraventricular\_tumor; pediatric\_posterior\_fossa\_tumor (diagnosis includes, but not limited to, medulloblastoma, ependymoma, pilocytic astrocytoma); brainstem\_glioma (diagnosis also called diffuse midline gliomas); pituitary\_tumor; rathkes\_cleft\_cyst; craniopharyngioma; chordoma\_chondrosarcoma; chiari\_malformation (``yes'' for chiari I/II malformations); cerebral\_atrophy; brain\_herniation; brain\_mass\_effect; brain\_midline\_shift; cephaloceles; encephalomalacia\_gliosis; cerebral\_edema; cavum\_septum\_pellucidum; dandy\_walker\_malformation; mega\_cisterna\_magna; heterotopia; pachygyria\_lissencephaly; schizencephaly; dysgenesis\_corpus\_callosum; ventriculomegaly (``yes'' for any abnormally enlarged ventricles); hydrocephalus\_ex\_vacuo (``yes'' for ventriculomegaly due to brain atrophy, normal pressure hydrocephalus, or ex vacuo dilation); obstructive\_hydrocephalus (``yes'' for ventriculomegaly due to mass lesion or aqueductal stenosis); subdural\_hygroma\_effusions; intracranial\_hypotension; craniotomy\_craniectomy; tumor\_resection\_cavity; postsurgical\_changes; pneumocephalus; catheter (``yes'' for any intracranial catheter); spine\_syrinx.
}
\end{figure*}

\clearpage

\begin{figure*}[p!]
    \centering\includegraphics[scale=0.75]{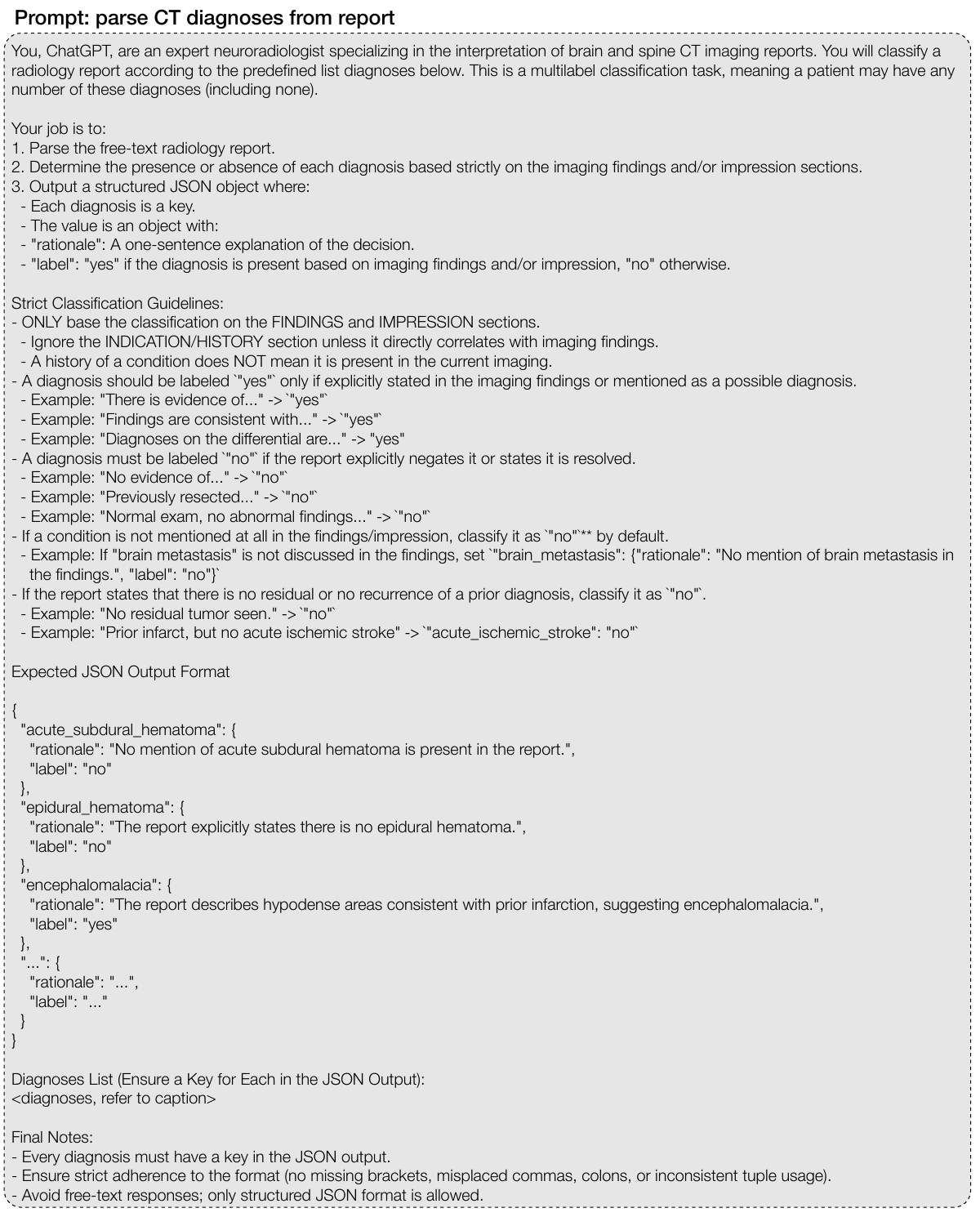}
    \caption{\textbf{CT diagnosis extraction prompt}. Caption on next page.}
    \label{exfig:supp_data3}
\end{figure*}

\begin{figure*}[p!]\ContinuedFloat
    \caption{\textbf{CT diagnosis extraction prompt.} We used an LLM-based annotation pipeline \cite{Lyu2025-bt} with GPT-4.1-mini to convert free-text radiology reports into structured labels for 82 expert-defined CT diagnoses. The prompt requires, for each diagnosis, a present/absent label and a brief rationale supporting the decision. The full list of CT diagnoses is here: acute\_subdural\_hematoma (``yes'' if acute blood in subdural space, often after head trauma and associated with midline shift); subacute\_chronic\_subdural\_hematoma (``yes'' if subacute or chronic blood in the subdural space, often non-traumatic); epidural\_hematoma; brain\_contusion; displaced\_skull\_fracture (``yes'' if displaced or depressed skull fracture); nondisplaced\_skull\_fracture; skull\_base\_fracture; orbital\_trauma (``yes'' if orbit fracture, globe rupture, or other orbital trauma); orbital\_emphysema; pneumocephalus; diffuse\_axonal\_injury; craniofacial\_injury (``yes'' if any mandibular, maxillary, zygoma, or other facial fractures or trauma); intracranial\_hemorrhage (``yes'' any intracranial hemorrhage); intraparenchymal\_hemorrhage (``yes'' for hemorrhages within the brain, usually spontaneous and due to hypertension); aneurysmal\_subarachnoid\_hemorrhage (``yes'' if due to ruptured aneurysm or other vascular cause); traumatic\_subarachnoid\_hemorrhage (``yes'' if due to head trauma); intraventricular\_hemorrhage; scalp\_hemorrhage\_hematoma; acute\_ischemic\_stroke (``yes'' for acute ischemia/infarct often with cerebral edema, mass effect, midline shift); small\_vessel\_ischemic\_disease (also known as microvascular ischemic disease); large\_vessel\_occlusion (``yes'' if large-vessel occlusion found on CT angiogram); intracranial\_atherosclerosis; arterial\_dissection (``yes'' if dissection found on CT angiography); carotid\_cavernous\_fistula; cerebral\_venous\_sinus\_thrombosis (``yes'' if thrombosis found on CT angiogram/venogram); intracranial\_aneurysm (``yes'' if found on CT angiogram); cavernous\_malformation\_cavernoma (``yes'' if radiologist confident for cavernoma; ``no'' if possible microbleed); cerebral\_arteriovenous\_malformation; intra\_axial\_brain\_tumor (``yes'' for tumors within the brain parenchyma); extra\_axial\_brain\_tumor (``yes'' for tumors outside the brain parenchyma, but inside the skull/intracranial space); orbital\_tumor (``yes'' if diagnosis of any orbital mass, growth, or neoplasm); pineal\_tumor (``yes'' for any tumor in the pineal region); pituitary\_tumor; skull\_bone\_tumor; posterior\_fossa\_tumors (``yes'' for any tumor in the posterior fossa, cerebellum, or brainstem); head\_neck\_tumor (``yes'' if diagnosis of tumor/cancer in the oral cavity, throat, larynx, nasal cavity, salivary, and thyroid gland); spine\_tumor (``yes'' if diagnosis of any spine mass, growth, or neoplasm); spinal\_degenerative\_changes; arachnoid\_cyst; intracranial\_epidermoid\_dermoid\_cyst; colloid\_cyst; cerebral\_subdural\_empyema; cerebral\_abscess; craniotomy; cranioplasty\_implant (``yes'' if artificial implant to repair skull defect); burr\_hole; resection\_cavity; postsurgical\_changes; catheter (``yes'' for any intracranial catheter); intracranial\_pressure\_monitor; brain\_lead (``yes'' for brain electrode leads used in functional neurosurgery); aneurysm\_coil; aneurysm\_clip; embolization\_material; cerebral\_foreign\_body (examples include glass, metal, plastic, wood); ventriculomegaly; obstructive\_hydrocephalus; transependymal\_flow; subdural\_hygroma\_effusions; slit\_ventricle; intracranial\_hypotension; cerebral\_edema; cerebral\_atrophy; brain\_herniation; brain\_mass\_effect; midline\_shift; encephalomalacia\_gliosis; basal\_ganglia\_calcification; chiari\_malformation (``yes'' for Chiari I/II malformations); dysgenesis\_corpus\_callosum; dandy\_walker\_malformation; mega\_cisterna\_magna; sinusitis; mastoid\_effusion (includes mastoid fluid and mastoiditis); otitis\_media; peritonsillar\_abscess; pharyngeal\_abscess (includes retropharyngeal and parapharyngeal abscesses); cholesteatoma; head\_neck\_enlarged\_lymph\_node; airway\_obstruction (examples include septal deviation, choanal atresia, obstruction, airway edema, laryngeal edema, and pharyngeal edema); thyroid\_nodule; parathyroid\_nodule.
}
\end{figure*}

\clearpage

\begin{figure*}[p!]
    \centering\includegraphics[scale=0.75]{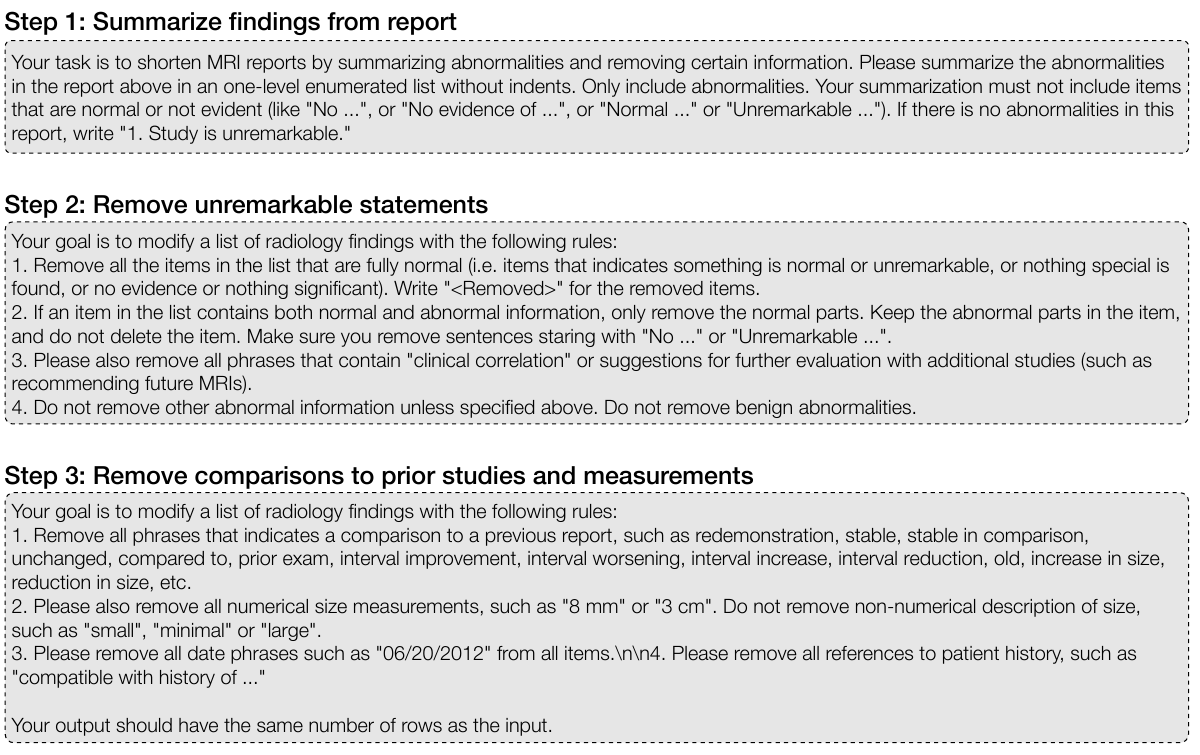}
    \caption{\textbf{Prompts for structured findings extraction.} Three successive prompts were applied to GPT-4.1-mini to turn free-text radiology reports into an itemized list of structured findings for report generation. The pipeline extracts key positive findings and removes unremarkable statements, comparisons to prior studies, and raw measurements. The resulting findings supervised training of NeuroVFM-LLaVA for preliminary findings generation.}
    \label{exfig:supp_data4}
\end{figure*}
\clearpage

\begin{figure*}[p!]
    \centering\includegraphics[scale=0.75]{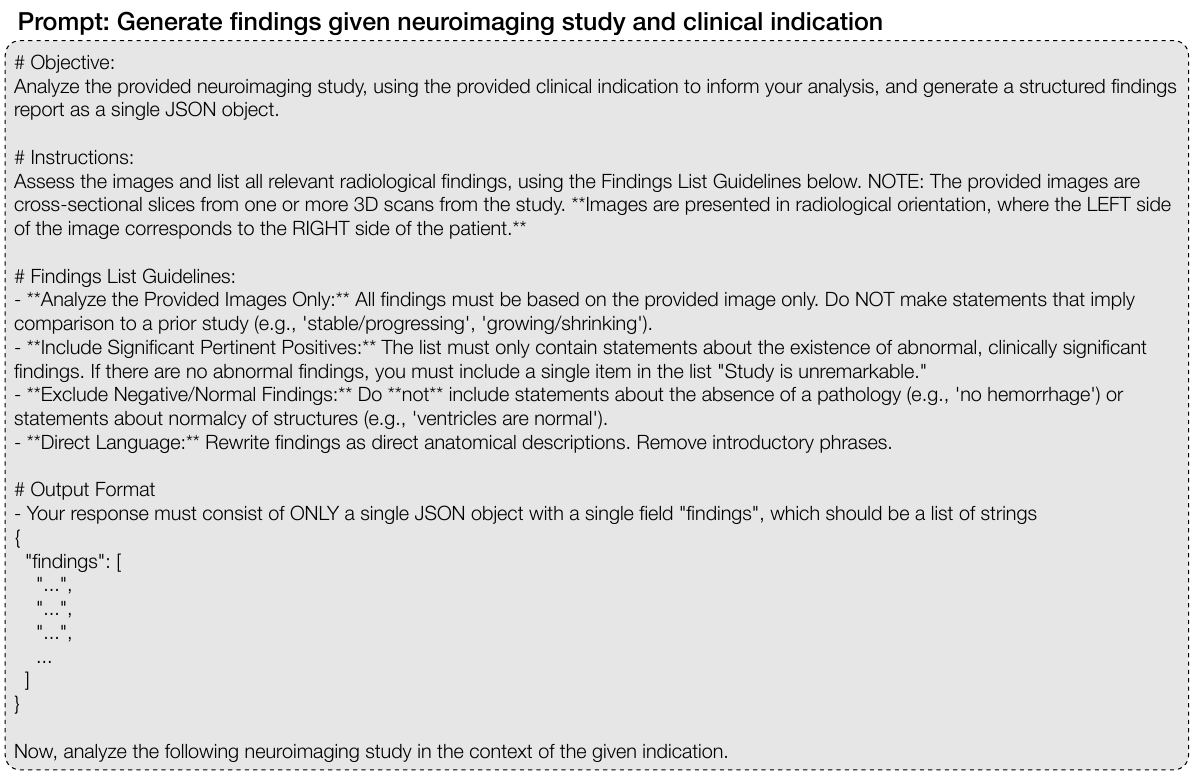}
    \caption{\textbf{Frontier model prompt for study findings.} To assess reasoning models (e.g., GPT-5-thinking) on preliminary neuroimaging findings generation, we passed the clinical indication and the study slice-by-slice. The prompt asks for an itemized list of key findings at the study level.}
    \label{exfig:supp_data5}
\end{figure*}
\clearpage

\begin{figure*}[p!]
    \centering\includegraphics[scale=0.75]{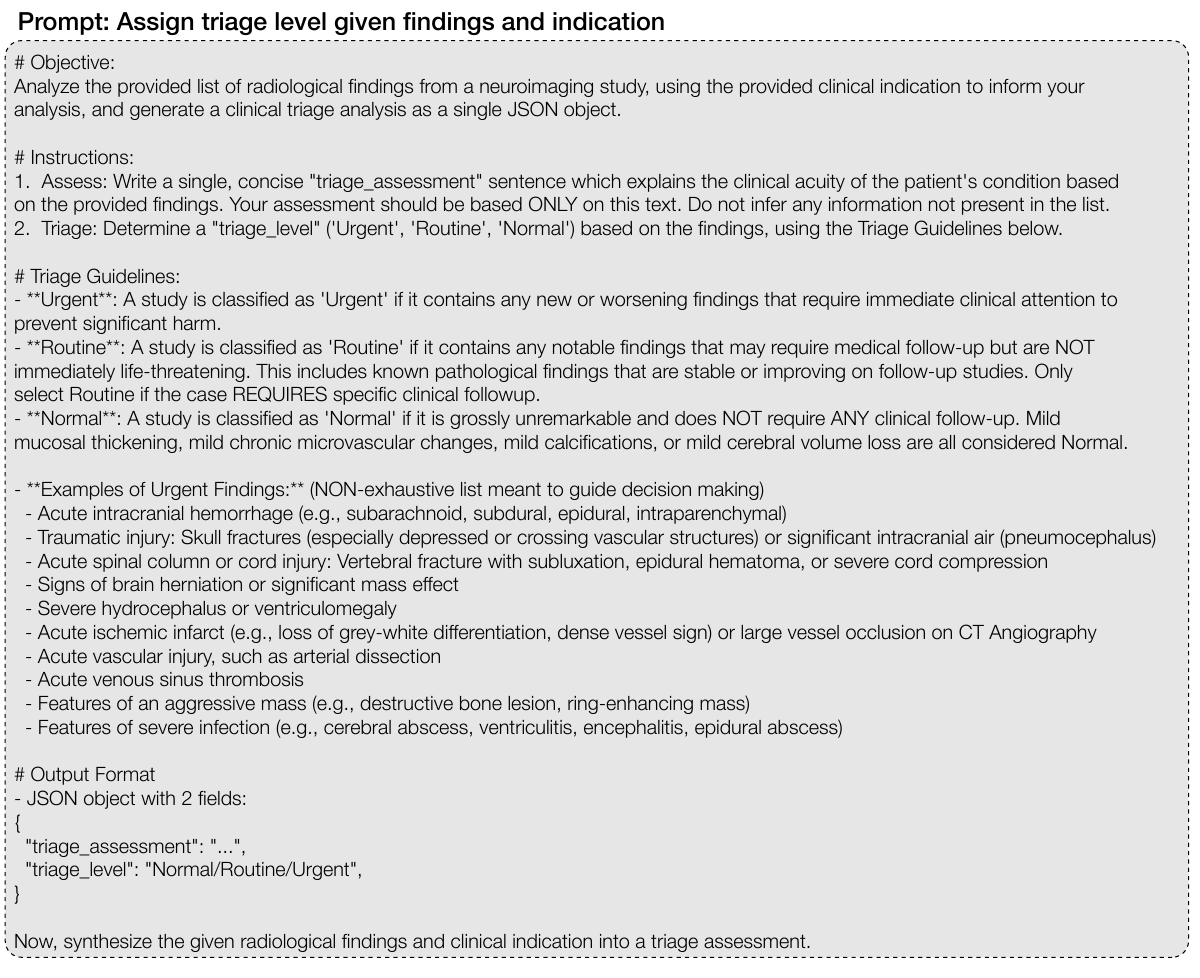}
    \caption{\textbf{Triage prompt given findings and indication.} Using a standardized protocol, we supply the clinical indication and itemized findings to a reasoning model (e.g., GPT-5-thinking) and request a single triage label with a brief justification. Labels are drawn from an expert-defined schema aligned with routine clinical triage. When provided ground-truth reports, both models show high accuracy and close agreement with clinician labels (Extended Data Fig. \ref{exfig:ex_data9}b), supporting the validity of their reasoning.}
    \label{exfig:supp_data6}
\end{figure*}
\clearpage

\renewcommand{\figurename}{Supplementary Data Table}
\setcounter{figure}{0}

\begin{figure*}[p!]
    \caption{\textbf{Descriptive characteristics of the UM-NeuroImages prospective test set.} The temporally held-out UM-NeuroImages test set comprises 50,293 CT and MRI studies. For each study, the table reports modality, scanner manufacturer, acquisition site, patient age at scan (with ages $\geq$90 years truncated to 90 years due to PHI constraints), sex, ethnicity, and MRI field strength (when applicable). \textit{This large table is available in the accompanying spreadsheet.}}
    \label{exfig:supp_table1}
\end{figure*}

\begin{figure*}[p!]
    \caption{\textbf{Full task-level performance on the UM-NeuroImages CT test set.} The temporally held-out UM-NeuroImages CT test set comprises 21,054 CT studies with 82 study-level CT diagnoses. For each diagnosis and diagnostic category, the table reports the classification performance of NeuroVFM and three baseline models (HLIP, DINOv3, and BiomedCLIP). \textit{This large table is available in the accompanying spreadsheet.}}
    \label{exfig:supp_table2}
\end{figure*}

\begin{figure*}[p!]
    \caption{\textbf{Full task-level performance on the UM-NeuroImages MRI test set.} The temporally held-out UM-NeuroImages MRI test set comprises 29,239 MRI studies with 74 study-level MRI diagnoses. For each diagnosis and diagnostic category, the table reports the classification performance of NeuroVFM and three baseline models (HLIP, DINOv3, and BiomedCLIP). \textit{This large table is available in the accompanying spreadsheet.}}
    \label{exfig:supp_table3}
\end{figure*}

\begin{figure*}[p!]
    \caption{\textbf{Performance of NeuroVFM on external neuroimaging benchmarks.} The table summarizes the performance of NeuroVFM and three baseline models (HLIP, DINOv3, and BiomedCLIP) on eight external neuroimaging benchmarks (six MRI and two CT datasets). All models were evaluated using 8-fold stratified cross-validation with a 20\% held-out test split, except for AIBL and OASIS-1, which were evaluated on the entire dataset using the ADNI-trained classifier. \textit{This large table is available in the accompanying spreadsheet.}}
    \label{exfig:supp_table4}
\end{figure*}

\begin{figure*}[p!]
    \caption{\textbf{Generated reports and evaluations for the UM-NeuroImages-Triage test set.} For the balanced UM-NeuroImages-Triage test set of 300 studies, the table includes generated reports from all models, the corresponding ground-truth radiology reports, clinical indication, triage labels, and all associated evaluation scores from expert raters. \textit{This large table is available in the accompanying spreadsheet.}}
    \label{exfig:supp_table5}
\end{figure*}

\begin{figure*}[p!]
    \caption{\textbf{Hyperparameters for NeuroVFM pretraining and UM-NeuroImages study-level attentive probing.} The table lists all hyperparameters used to (1) pretrain the 3D ViT backbone and (2) train the study-level attentive probe used for UM-NeuroImages. For each configuration, we report optimization settings, data sampling strategies, architectural choices, and regularization parameters. A similar probing strategy was used for all external benchmarks, with hyperparameters adjusted for dataset size and complexity. \textit{This large table is available in the accompanying spreadsheet.}}
    \label{exfig:supp_table6}
\end{figure*}

\begin{figure*}[p!]
    \caption{\textbf{Hyperparameters for NeuroVFM-LLaVA training and inference.} The table lists all hyperparameters used to perform LLaVA-1.5-style vision-instruction tuning on top of the frozen NeuroVFM backbone to finetune a language model for key findings generation. Reported settings include model initialization, optimizer configuration, learning rate schedule, batch size, context length, training duration, and inference parameters. The deployed inference pipeline is implemented with HuggingFace to enable standardized report generation. \textit{This large table is available in the accompanying spreadsheet.}}
    \label{exfig:supp_table7}
\end{figure*}